\documentclass[10pt, twocolumn, a4paper, logo, copyright]{googledeepmind}

\usepackage[utf8]{inputenc} 
\usepackage[T1]{fontenc}    
\usepackage{url}            
\usepackage{booktabs}       
\usepackage{amsfonts}       
\usepackage{nicefrac}       
\usepackage{microtype}      
\usepackage{xcolor}         

\usepackage{commands}
\usepackage{amsmath}
\usepackage{algorithm}
\usepackage{algpseudocode}
\usepackage{url}
\usepackage{xr}
\usepackage{listings}
\usepackage{adjustbox}
\usepackage{subcaption}
\usepackage{xspace}
\usepackage{longtable}
\usepackage{caption}
\usepackage{tcolorbox}
\usepackage{xskak}
\usepackage{multicol}
\usepackage{multirow}
\usepackage{wrapfig}
\usepackage{makecell}
\usepackage{rotating}

\usepackage{commands}
\usepackage{amsmath}
\usepackage{amsthm}
\usepackage{algorithm}
\usepackage{algpseudocode}
\usepackage{xr}
\usepackage{xcolor}
\usepackage{array}
\usepackage{enumitem}
\usepackage{minted}
\usepackage[authoryear, sort&compress, round]{natbib}
\usepackage{comment}

\usepackage[colorlinks=true]{hyperref}

\theoremstyle{definition}
\newtheorem{definition}{Definition}[section]

\usepackage[capitalize]{cleveref}

\usepackage[authoryear, sort&compress, round]{natbib}
\defcitealias{blanchard2026determination}{The bbchallenge Collaboration, 2026}

\renewcommand{\today}{Oct 15th, 2026} 

\usepackage{subfiles} 

\usepackage{subfiles} 

\title{At the Limits of Computability: Steps Towards Autonomous AI Discovery on the Busy Beaver Frontier}

\author{Marco Dos Santos$^{1,2}$, Vivek Veeriah$^{1}$, Alex Havrilla$^{1}$, Andrew Bolt$^{1}$, Goran Zuzic$^{1}$, Satinder Singh$^{1}$ and Tom Zahavy$^{1}$. $^{1}$ Google DeepMind, $^2$ University of Cambridge. Work done at Google DeepMind.}

\begin{abstract}
Recent breakthroughs in AI for mathematics—ranging from Olympiad medal performances to resolving long-standing open conjectures—have rapidly expanded what automated systems can achieve. Yet these successes primarily concern predefined, human-formulated problems, leaving open how AI systems can autonomously uncover mathematical structure and formulate new questions. To push beyond static challenges, we explore the Busy Beaver (BB) domain, an uncomputable frontier that is fundamentally open-ended. Because no algorithm can compute the Busy Beaver number $\mathrm{BB}(n)$ for every $n$, this domain provides a natural setting for sustained exploration through mathematical conjecture and proof. In this paper, we introduce a Gemini-based agentic pipeline that automatically discovers accelerations of Turing machines, expressed as compact mathematical rules summarizing many execution steps. Our system integrates pattern recognition, automated conjecturing, informal mathematical reasoning, and formal theorem proving in Lean. With this framework, our system autonomously discovered and formally verified novel Collatz-like accelerations for eleven machines in $\mathrm{BB}(6)$. Subject to community acceptance, we provisionally classify these machines as \textit{Cryptids}, namely machines whose blank-tape executions admit exact, compact mathematical descriptions but whose halting behavior remains unresolved. These results advance the analysis of $\mathrm{BB}(6)$ holdouts on the Busy Beaver frontier and demonstrate the potential of autonomous AI systems for open-ended mathematical discovery in fundamental computer science.
\end{abstract}

\begin{document}
\maketitle

\section{Introduction}

Recent milestones in AI-driven mathematics—from an AI system earning an IMO medal \citep{hubert2026olympiad} to formalizing \href{https://github.com/anthropics/fermats-last-theorem}{Fermat's Last Theorem in Lean 4} or to OpenAI's multi-agent system proposing a solution to the Navier-Stokes existence and smoothness problem \citep{OpenAI2026NavierStokes}—have largely focused on using AI to \textit{solve} pre-defined, human-formulated questions. Gowers' distinction between problem-solving and theory-building \citep{gowers2000two}, together with the emphasis on mathematical understanding in the \href{https://leidendeclaration.ai/}{Leiden Declaration} and Tao's \textit{Mathematics in the Age of AI} \citep{tao2026mathematics}, motivates a broader goal for mathematical AI: contributing to the discovery of mathematical structures and the formulation of new questions. Motivated by this broader goal, our work explores a complementary paradigm: using AI to autonomously uncover and formalize mathematical problems arising from an open-ended stream of computation. The ultimate testbed for this kind of open-ended discovery lies at the  frontier of computability theory: the Busy Beaver Challenge.

Introduced by \citet{rado1962non}, the challenge poses a simple question: among all halting Turing Machines (TMs) of a given size, what is the maximum number of steps a machine can run before halting? For machines with $n$ states and two symbols, started on a blank tape, this maximum is denoted by $\mathrm{BB}(n)$. Knowing the value of $\mathrm{BB}(n)$ is powerful because it provides a universal halting criterion for all $n$-state machines: if a machine runs for $\mathrm{BB}(n) + 1$ steps, it is guaranteed to run forever. Consequently, the function $n \mapsto \mathrm{BB}(n)$ is uncomputable; otherwise, the Halting Problem would be decidable. Moreover, the sequence of Busy Beaver numbers—$\mathrm{BB}(1), \mathrm{BB}(2), \dots$—grows asymptotically faster than any computable sequence, including stacked exponentials, the Ackermann function, and $\text{TREE}(n)$ \citep{aaronson2020busy}. 

An intriguing consequence of the uncomputability of the Busy Beaver numbers is that mathematical creativity is inherently open-ended \citep{chaitin2013proving}. Its significance extends beyond computation: statements such as Fermat's Last Theorem, the Riemann hypothesis, Goldbach's conjecture, and the consistency of ZF can be encoded as the non-halting of specific Turing machines \citep{aaronson2020busy}. Because no algorithm can compute $\mathrm{BB}(n)$ for all $n$, progress along this frontier offers continuing scope for mathematical conjecture, proof, and the development of new techniques. This makes it a great testing ground for autonomous AI discovery. Regardless of how sophisticated future
\clearpage
\begin{table*}[h]
\centering
\scriptsize 
\begin{tabularx}{\textwidth}{l c c X}
\toprule
\textbf{Machine} & \makecell[c]{\textbf{BB} \\ \textbf{domain}} & \makecell[c]{\textbf{Discovery} \\ \textbf{type}} & \textbf{Comment} \\
\midrule
\multicolumn{4}{l}{\textbf{Part I: New Discoveries}} \\
\midrule

\texttt{1RB1LD\_\allowbreak 1RC0RA\_\allowbreak 1LA1RE\_\allowbreak 0LA0LB\_\allowbreak 0RB1RF\_\allowbreak -{}-{}-1RD} & $\mathrm{BB}(6)$ & Full  & - \\ \addlinespace
\texttt{1RB1RD\_\allowbreak 0LC1RA\_\allowbreak 0RA1LC\_\allowbreak 1RA1LE\_\allowbreak 1RF0LD\_\allowbreak -{}-{}-0RE} & $\mathrm{BB}(6)$ & Full  & $\dagger$. \\ \addlinespace
\texttt{1RB1RB\_\allowbreak 1LC0RE\_\allowbreak -{}-{}-1LD\_\allowbreak 0RA1LB\_\allowbreak 1RA1LF\_\allowbreak 0LE1LE} & $\mathrm{BB}(6)$ & Full  & $\dagger$ \\ \addlinespace
\texttt{1RB0LF\_\allowbreak 1LC1RA\_\allowbreak 0RE0RD\_\allowbreak -{}-{}-1LE\_\allowbreak 1LF1RC\_\allowbreak 1LC1LA} & $\mathrm{BB}(6)$ & Full &  $\dagger$ \\ \addlinespace
\texttt{1RB1RD\_\allowbreak 0LC1RA\_\allowbreak 1RA1LC\_\allowbreak 0RB1LE\_\allowbreak 1RF0LD\_\allowbreak -{}-{}-0RE} & $\mathrm{BB}(6)$ & Full  & $^{\dagger\star}$ \\ \addlinespace
\texttt{1RB0LF\_\allowbreak 1RC1RA\_\allowbreak 0RD0RB\_\allowbreak 1LE-{}-{}-\_\allowbreak 0LA1LE\_\allowbreak 1RD1LA} & $\mathrm{BB}(6)$ & Full  & $^{\star}$ \\ \addlinespace
\texttt{1RB0LC\_\allowbreak 0LC1RF\_\allowbreak 1LE1RD\_\allowbreak 0RC1RC\_\allowbreak 0LA1LA\_\allowbreak -{}-{}-1RA} & $\mathrm{BB}(6)$ & Full  & $^{\dagger\star}$ \\ \addlinespace
\texttt{1RB1LF\_\allowbreak 0RC1RD\_\allowbreak 1LA1LD\_\allowbreak 1LE0RA\_\allowbreak -{}-{}-1LC\_\allowbreak 0LA1LA} & $\mathrm{BB}(6)$ & Full  & $^{\dagger\star}$ \\ \addlinespace
\texttt{1RB1RC\_\allowbreak 1LC1LC\_\allowbreak 1RE0LD\_\allowbreak 1LB1RF\_\allowbreak -{}-{}-1RA\_\allowbreak 0RD1RD} & $\mathrm{BB}(6)$ & Full  & $^{\dagger\star}$ \\ \addlinespace
\texttt{1RB1LE\_\allowbreak 1RC1RA\_\allowbreak 0LD1RB\_\allowbreak 1LE1LD\_\allowbreak 1RF0LA\_\allowbreak -{}-{}-0RE} & $\mathrm{BB}(6)$ & Full  & $^{\dagger\star}$ \\ \addlinespace
\texttt{1RB1LA\_\allowbreak 1LC0RE\_\allowbreak 1LF1LD\_\allowbreak 1RE0LC\_\allowbreak 1LB1RA\_\allowbreak -{}-{}-0LA} & $\mathrm{BB}(6)$ & Full  & Mentioned on the \href{https://wiki.bbchallenge.org/wiki/User:RobinCodes/Next_Steps}{Wiki}, with no conjectured rules.$^{\dagger\star}$ \\ \addlinespace
\midrule
\multicolumn{4}{l}{\textbf{Part II: Rediscoveries}} \\
\midrule

\texttt{1RB3LA1LA1RA1RA\_\allowbreak 2LB2RA-{}-{}-4RB1LB} & $\mathrm{BB}(2,5)$ & Acceleration  & First analyzed by mxdys on 12 Mar 2025, appears to be a Cryptid. (\href{https://wiki.bbchallenge.org/wiki/1RB3LA1LA1RA1RA_2LB2RA---4RB1LB}{Wiki}) \\ \addlinespace
\texttt{1RB0RB\_\allowbreak 1LB0LC\_\allowbreak 1RD1LC\_\allowbreak 1RE0RA\_\allowbreak 0RF1RA\_\allowbreak -{}-{}-1LB} & $\mathrm{BB}(6)$ & Proof  & The result is similar to a \href{https://discord.com/channels/960643023006490684/1239205785913790465/1496521465980518431}{comment on Discord} by the user Sheep. We provide proofs. \\ \addlinespace
\texttt{1RB-{}-{}-\_\allowbreak 0RC0RD\_\allowbreak 1LD1RB\_\allowbreak 0LE0LC\_\allowbreak 1RA0LF\_\allowbreak 1LD1LE} & $\mathrm{BB}(6)$ & Proof  & Analysis by Shawn Ligocki (\href{https://wiki.bbchallenge.org/wiki/1RB---_0RC0RD_1LD1RB_0LE0LC_1RA0LF_1LD1LE}{Wiki}). \\ \addlinespace
\texttt{1RB1RF\_\allowbreak 0LC0RF\_\allowbreak 1RD1LC\_\allowbreak -{}-{}-0LE\_\allowbreak 0RC1LF\_\allowbreak 1RA0LE} & $\mathrm{BB}(6)$ & Proof  & The result is similar to a \href{https://discord.com/channels/960643023006490684/1239205785913790465/1431334486649471078}{comment on Discord} by Shawn Ligocki. We provide proofs. \\ 
\addlinespace
\texttt{1RB0LE\_\allowbreak 1LC1RA\_\allowbreak 1RA1LD\_\allowbreak 0LC0LA\_\allowbreak 0RE1RF\_\allowbreak 0RC-{}-{}-} & $\mathrm{BB}(6)$ & Proof  & Mentioned on Discord with conjectured rules. We provide proofs.$^{\ddagger\star}$ \\ \addlinespace
\texttt{1RB1RE\_\allowbreak 1LC0RA\_\allowbreak 1RD0LB\_\allowbreak 1LB1RC\_\allowbreak 1LF0RD\_\allowbreak -{}-{}-0LE} & $\mathrm{BB}(6)$ & Proof  & BMO problem 5 (BMO5, TNF from state D; equivalent to Suspect~4); analysis by mxdys (\href{https://wiki.bbchallenge.org/wiki/1RB1RE_1LC0RA_1RD0LB_1LB1RC_1LF0RD_---0LE}{Wiki}). We provide proofs.$^{\ddagger\star}$ \\ \addlinespace
\texttt{1RB0LD\_\allowbreak 1LC0RA\_\allowbreak 1RA1LB\_\allowbreak 1LA1LE\_\allowbreak 1RF0LC\_\allowbreak -{}-{}-0RE} & $\mathrm{BB}(6)$ & Proof  & BMO problem 5 (BMO5, TNF from state A; equivalent to Suspect~13); analysis by mxdys (\href{https://wiki.bbchallenge.org/wiki/1RB0LD_1LC0RA_1RA1LB_1LA1LE_1RF0LC_---0RE}{Wiki}). We provide proofs.$^{\ddagger\star}$ \\ \addlinespace
\texttt{1RB0RF\_\allowbreak 1RC1RF\_\allowbreak 1LD0LE\_\allowbreak -{}-{}-1LC\_\allowbreak 1RA1LE\_\allowbreak 1LC1RB} & $\mathrm{BB}(6)$ & Proof  & Rules conjectured on Discord; similar to Space Needle (\href{https://wiki.bbchallenge.org/wiki/1RB0RF_1RC1RF_1LD0LE_---1LC_1RA1LE_1LC1RB}{Wiki}). We provide proofs.$^{\ddagger\star}$ \\ \addlinespace
\texttt{1RB0LD\_\allowbreak 0RC1RB\_\allowbreak 0RD0RA\_\allowbreak 1LE0RD\_\allowbreak 1LF-{}-{}-\_\allowbreak 0LA1LA} & $\mathrm{BB}(6)$ & Proof  & BMO problem 8 (BMO8); analysis by mxdys (\href{https://wiki.bbchallenge.org/wiki/1RB0LD_0RC1RB_0RD0RA_1LE0RD_1LF---_0LA1LA}{Wiki}). We provide proofs.$^{\ddagger\star}$ \\ \addlinespace
\texttt{1RB1LC\_\allowbreak 1LC1RD\_\allowbreak 1LA0LB\_\allowbreak 0RB1RE\_\allowbreak 1RF0RA\_\allowbreak -{}-{}-0RC} & $\mathrm{BB}(6)$ & Proof  & Tracked as Suspect~46 and Old Suspect~02; analysis by Racheline (\href{https://wiki.bbchallenge.org/wiki/1RB1LC_1LC1RD_1LA0LB_0RB1RE_1RF0RA_---0RC}{Wiki}). We provide proofs.$^{\ddagger\star}$ \\ \addlinespace
\texttt{1RB1LE\_\allowbreak 1LC0RA\_\allowbreak 0RF0LD\_\allowbreak 1LE1LA\_\allowbreak 1RC0LB\_\allowbreak -{}-{}-1RC} & $\mathrm{BB}(6)$ & Proof  & Rules by the user -d on \href{https://discord.com/channels/960643023006490684/1239205785913790465/1411637417709932544}{Discord} (\href{https://wiki.bbchallenge.org/wiki/1RB1LE_1LC0RA_0RF0LD_1LE1LA_1RC0LB_---1RC}{Wiki}). We provide proofs.$^{\ddagger\star}$ \\ \addlinespace
\texttt{1RB1LA\_\allowbreak 1RC1RE\_\allowbreak 1LD0RB\_\allowbreak 1LA0LC\_\allowbreak 0RF0RD\_\allowbreak 0RB-{}-{}-} & $\mathrm{BB}(6)$ & Proof  & Analysis by mxdys (\href{https://wiki.bbchallenge.org/wiki/1RB1LA_1RC1RE_1LD0RB_1LA0LC_0RF0RD_0RB---}{Wiki}); the community formalized the rules in Lean, with non-halting conditional on the Baker--W\"ustholz theorem. We provide proofs.$^{\ddagger\star}$ \\ \addlinespace
\texttt{1RB1RE\_\allowbreak 1LC0RA\_\allowbreak 0RD1LB\_\allowbreak -{}-{}-1RC\_\allowbreak 1LF1RE\_\allowbreak 0LB0LE} & $\mathrm{BB}(6)$ & Proof  & BMO1, discovered by the user -d (\href{https://wiki.bbchallenge.org/wiki/1RB1RE_1LC0RA_0RD1LB_---1RC_1LF1RE_0LB0LE}{Wiki}); its rules were previously proved in Lean. We provide proofs.$^{\ddagger\star}$ \\ \addlinespace
\texttt{1RB1RF\_\allowbreak 1LC1LF\_\allowbreak 0RE1LD\_\allowbreak 0LB1LD\_\allowbreak -{}-{}-1RC\_\allowbreak 1RA0RD} & $\mathrm{BB}(6)$ & Proof  & Found by Racheline (\href{https://wiki.bbchallenge.org/wiki/1RB1RF_1LC1LF_0RE1LD_0LB1LD_---1RC_1RA0RD}{Wiki}); rules conjectured on Discord. We provide proofs.$^{\ddagger\star}$ \\ \addlinespace
\texttt{1RB1LC\_\allowbreak 1RC0LD\_\allowbreak 1LA0RB\_\allowbreak 1LB1LE\_\allowbreak 1RF0LA\_\allowbreak -{}-{}-0RE} & $\mathrm{BB}(6)$ & Proof  & Analysis by mxdys (\href{https://wiki.bbchallenge.org/wiki/1RB1LC_1RC0LD_1LA0RB_1LB1LE_1RF0LA_---0RE}{Wiki}). BMO5 started in state C; equivalent to the two BMO5 machines above. We provide proofs.$^{\ddagger\star}$ \\
\bottomrule
\end{tabularx}
\caption{Collatz-like accelerations discovered by our system. We distinguish three discovery types. Full: new acceleration and proof. Acceleration: new acceleration for a known machine, with its proof. Proof: formal proof of rules similar to known ones. Appendix~\ref{app:rediscoveries} provides a detailed comparison between our rule systems and those previously proposed by the community. For machines with known or suspected accelerations, we cite the original source and highlight the added value of our analysis, such as a formal proof or a higher-level acceleration. For previously unanalyzed machines, our results represent complete, novel discoveries. We label machines with $\dagger$ if they were mentioned in a \href{https://discord.com/channels/960643023006490684/1239205785913790465/1422696822945743031}{Discord message by dyuan01} as holdouts with a high "complexity score", but with no rules or analysis, and with $\ddagger$ rediscoveries that were formally proved by our system but we did not check by hand. We label machines with $\star$ if they were discovered in a second stage of our pipeline. Specifically, following the completion of the full pipeline, we used a general-purpose auto-research system equipped with the recent Gemini 3.8 Flash model, which had access to all the results and code in this paper. We tasked it to conjecture and prove the equivalence of all the remaining suspects. This system was granted the ability to selectively make progress on machines and was incentivized to find as many accelerations as possible, with more compressed accelerations earning higher point values. The prompt used for this second stage is provided in Appendix~\ref{app:prompt_proving}.}
\label{tab:merged-cryptid-results}
\end{table*}
\clearpage

AI systems become, the Busy Beaver frontier will continue to present problems beyond the reach of any given algorithm.

In practice, searching for Busy Beaver values involves applying \textit{deciders} to vast spaces of Turing machines, progressively reducing them to smaller sets of \textit{holdouts} whose halting status remains unknown. For the purposes of this study, we focus on a specific subset of approximately $1{,}000$ holdouts in $\mathrm{BB}(6)$, detailed in \cref{table:BB}. Resolving a holdout ultimately requires proving that the corresponding machine either halts or runs forever. For some holdouts, however, mathematical analysis reveals dynamics whose long-term behavior corresponds to an open mathematical problem. Such machines are referred to as \textit{Cryptids} (\cref{def:cryptid}), and many known examples are governed by Collatz-like systems. While the fifth Busy Beaver number, $\mathrm{BB}(5)$, was recently determined through a collaborative community effort \citepalias{blanchard2026determination}, the next frontiers—$\mathrm{BB}(6)$, $\mathrm{BB}(2,5)$, and $\mathrm{BB}(3,3)$—contain Cryptids whose unresolved halting behavior prevents their complete classification. The Collatz conjecture itself remains a notoriously difficult open problem in mathematics, leading Paul Erd{\H{o}}s to remark that ``Mathematics is not yet ready for such problems.'' Conway's proof that a generalized Collatz reachability problem is algorithmically undecidable \citep{conway1972unpredictable} motivates a broader question: could some Cryptids, despite their short descriptions, lie at the edge of what ZFC can prove? \citet{aaronson2020busy} explores this question, conjecturing that ZF cannot determine the value of $\mathrm{BB}(20)$. Whether the Cryptids encountered in $\mathrm{BB}(6)$ approach such a frontier of provability remains speculative. To make progress despite these unresolved halting problems, the research community has shifted focus toward identifying Cryptids and determining the Busy Beaver numbers strictly among non-Cryptid candidates.
Against this background, our work focuses on accelerating machine execution. An acceleration summarizes many low-level Turing machine steps through compact transition rules, enabling substantially faster simulation and exposing the machine's higher-level arithmetic structure. Such an acceleration may facilitate the analysis of halting behavior or reveal a mapping to a Collatz-like system.

To execute this strategy, we introduce an automated, Gemini-based \textbf{coding scientist} that compresses simulated execution data into structured mathematical equations and proves their equivalence in \href{https://lean-lang.org/}{Lean}. Rather than statically analyzing the specific transition rules of the Turing machines, we treat them as "grey boxes"—maintaining full visibility of their internal state during execution while remaining agnostic to their underlying code. Because it relies purely on observing simulated behavior, our approach remains general; in principle, it is applicable to any Turing-complete programming language like Python and C++.~\footnote{BB numbers can be searched for in any Turing-complete programming language by asking: "What is the largest natural number that can be generated by any halting program of length $n$ bits?" \citep{chaitin2013proving}.}

\cref{fig:pipeline_overview} presents our proposed method across four key stages. First, a custom compression algorithm performs \textit{pattern detection} on simulated data, parsing it into blocks of symbols. Next, during \textit{automated conjecturing}, a Large Language Model (LLM) acts as a mathematical explorer. It synthesizes programmatic transition rules from these empirical patterns, using an interactive simulation tool to navigate edge cases and refine the rules into Collatz-like transition systems. Conjectures that pass strict empirical filters enter an \textit{informal proving} phase, where the same LLM operates in an iterative generator-evaluator loop, drafting step-by-step mathematical proofs and refining them in response to separate correctness and format evaluations. Finally, during \textit{formal verification}, an autonomous agent interacts directly with the Lean 4 environment to translate and compile these proofs, mathematically guaranteeing the machine's long-term dynamics.

\begin{table*}[t]
\centering
\caption{The Busy Beaver domain (Aug 2026). The initial search space of $(2m(n+1))^{mn}$ Turing machines is reduced to a set of \href{https://wiki.bbchallenge.org/wiki/Holdouts_lists}{holdouts} via \href{https://wiki.bbchallenge.org/wiki/Tree_Normal_Form}{canonization} 
 and \href{https://wiki.bbchallenge.org/wiki/Decider}{deciders}. Our pipeline kicks in and provides candidates from the holdouts as well as verified accelerations.}
\vspace{0.5cm}
\begin{adjustbox}{width=0.8\textwidth}
\begin{tabular}{lcccccc}
\toprule
& & \multicolumn{2}{c}{\textbf{Community Analysis}}
& \multicolumn{3}{c}{\textbf{Our Pipeline}} \\
\cmidrule(lr){3-4}\cmidrule(lr){5-7}
\textbf{Domain}
& \textbf{Machines}
& \textbf{Holdouts}
& \textbf{Cryptids}
& \textbf{Candidates}
& \textbf{Rediscoveries}
& \textbf{New Cryptids} \\
\midrule
$\mathrm{BB}(6)$ & $2.32 \times 10^{17}$ & 1094 & 18 & 50 & 14 & 11\\
$\mathrm{BB}(2,5)$ & $5.9 \times 10^{14}$ & 65 & 2 & 2 & 1 & 0\\
$\mathrm{BB}(3,3)$ & $2.64 \times 10^{12}$ & 4 & 1 & 0 & 0 & 0\\
\bottomrule
\end{tabular}
\end{adjustbox}
\label{table:BB}
\end{table*}

\cref{table:BB} summarizes our results alongside the status of the BB frontier domains of $\mathrm{BB}(6)$, $\mathrm{BB}(2,5)$, and $\mathrm{BB}(3,3)$. During the conjecturing phase, our pipeline autonomously identified 52 candidate machines whose behavior appeared to be Collatz-like. The complete list is provided in Appendix~\ref{app:candidates}. The \textbf{Cryptids} column details how many of the holdout machines are considered to be Cryptids (\cref{def:cryptid}). It is important to note that there are machines that are suspected to be Cryptids by the community (for example, there exists preliminary analysis for them on the Discord channel), but have not been formally recognized as Cryptids in the wiki yet. Our main result concerns eleven BB(6) holdouts with no prior documented analysis. For each of these, our system discovered and proved that its blank-tape execution is exactly described by a Collatz-like system, so that the machine halts if and only if this system reaches a halting configuration. These are new, formally verified reductions of each machine's halting problem to an arithmetic question. Subject to community acceptance, we provisionally classify these machines as Cryptids (\textbf{New Cryptids} column). For fifteen further machines previously studied by the community, we provide formal proofs of similar rules or new accelerations.  Table~\ref{tab:merged-cryptid-results} details all these machines, and Appendix~\ref{app:discovered-accelerations} gives the full transition systems for the new discoveries.

\section{Background}
\subsection{Turing Machines and the Busy Beaver Problem}
\label{subsec:background_bb}

A Turing machine is a theoretical model of computation consisting of a finite set of states and an infinite one-dimensional tape divided into discrete cells. Each cell contains a symbol from a finite alphabet, typically $\{0, 1\}$. In the Busy Beaver setting, the machine is initialized on a blank tape, with every cell containing the symbol $0$. The machine operates via a read/write head that observes the current cell and, based on a deterministic transition function, writes a new symbol, moves one cell to the left or right, and transitions to a new state. Execution continues iteratively until the machine reaches a designated halt state, at which point computation terminates.

The Busy Beaver problem, introduced by Tibor Radó in 1962 \citep{rado1962non}, seeks to find the maximum number of steps that a halting Turing machine with $n$ states and an $m$-symbol alphabet can take on an initially blank tape before eventually halting. This maximum value is denoted by the Busy Beaver function, $\mathrm{BB}(n,m)$. When considering machines with a 2-symbol alphabet, this notation is conventionally simplified to $\mathrm{BB}(n)$ rather than $\mathrm{BB}(n, 2)$. Because determining whether an arbitrary Turing machine halts is fundamentally undecidable due to the Halting Problem, $\mathrm{BB}(n)$ is uncomputable. To establish that $\mathrm{BB}(n)=B$, one must exhibit an $n$-state machine that halts after exactly $B$ steps and prove that every other $n$-state machine either halts in at most $B$ steps or never halts. Machines whose halting status from the blank tape remains unknown are called \textit{holdouts}. Resolving $\mathrm{BB}(n)$ therefore requires resolving every holdout, by establishing either that it halts and determining its runtime, or that it runs forever. Over the decades, community efforts have successfully resolved the initial values of this function: $\mathrm{BB}(2) = 6$, $\mathrm{BB}(3) = 21$, $\mathrm{BB}(4) = 107$, and, most recently, $\mathrm{BB}(5) = 47,176,870$ \citepalias{blanchard2026determination}. This exhaustive classification process leaves a frontier of highly complex, unclassified holdout machines in larger domains, such as the remaining open cases in $\mathrm{BB}(6)$, $\mathrm{BB}(3,3)$, and $\mathrm{BB}(2,5)$. For broader background, we refer the reader to the \href{https://wiki.bbchallenge.org/wiki/Main_Page}{BB Wiki}, a recent overview in \href{https://www.quantamagazine.org/amateur-mathematicians-find-fifth-busy-beaver-turing-machine-20240702/}{Quanta Magazine}, or historical surveys such as \citep{michel2009busy}.

\subsection{Cryptids and Collatz-like Systems}
\label{subsec:background_cryptids}

\begin{definition}[Cryptid]
\label{def:cryptid}
In this work, a \textit{Cryptid} is a holdout machine whose blank-tape execution admits an exact, compact description as a mathematical dynamical system, but for which determining whether the resulting trajectory reaches a halting condition remains unresolved. Thus, unlike a holdout in general, a Cryptid comes with a high-level mathematical description that exposes the structure underlying its unresolved halting behavior.
\end{definition}

This definition is broader than the conventional meaning of the term within the Busy Beaver community, where a machine is recognized as a Cryptid only if its associated mathematical problem is considered genuinely open and difficult. Accordingly, our classification of newly identified machines as Cryptids remains subject to community acceptance.

The dynamics of Cryptids frequently take the form of \textit{Collatz-like} systems. The Collatz-like structure generally becomes apparent when tape configurations are expressed as sequences of blocks of identical symbols, whose variable lengths are represented by exponents. The machine's evolution between these parameterized configurations can then be described by exact macro-transitions acting on those exponents. The resulting transition system is typically piecewise affine: the update applied to the exponents depends on their congruence classes, for example on whether a given exponent is even or odd. This is the sense in which the system is Collatz-like: as in the famous $3x+1$ map, different multiplicative and additive updates are selected by modular conditions. Besides revealing the machine's high-level arithmetic structure, this abstraction accelerates simulation by replacing many individual Turing machine steps with a single macro-transition. However, simple transition rules do not imply predictable behavior. \citet{conway1972unpredictable} showed that, for a class of generalized Collatz maps, deciding whether the trajectory from a given initial integer ever reaches $1$ is algorithmically undecidable. Although this does not establish independence from ZFC for any particular Cryptid, it motivates asking whether such machines, despite their short descriptions, might lie at the edge of what ZFC can prove. For the Cryptids encountered in $\mathrm{BB}(6)$, proximity to this frontier is speculative. Irrespective of these foundational questions, determining whether their trajectories ever reach a halting configuration remains an open problem and a central obstacle to completing the corresponding Busy Beaver classifications.

A concrete example of a Cryptid is \textit{Antihydra}, a six-state, two-symbol Cryptid in the $\mathrm{BB}(6)$ domain. Consider the parameterized tape configurations
$$A(a,b):=0^\infty\ 1^a\ 0\ 1^b\ \mathrm{E}{>}\ 0^\infty,$$
where $\mathrm{E}{>}$ records the position of the head and its current state. Antihydra satisfies the following exact macro-transitions:
$$
\begin{aligned}
A(a,2b) &\xrightarrow{2a+3b^2+12b+11} A(a+2,3b+2),\\
A(0,2b+1) &\xrightarrow{3b^2+9b-1} 0^\infty\ {<}\mathrm{F}\ 110\ 1^{3b}\ 0^\infty,\\
A(a+1,2b+1) &\xrightarrow{3b^2+12b+5} A(a,3b+3).
\end{aligned}
$$
The labels above the arrows give the exact numbers of underlying Turing machine steps. Throughout, $a, b \geq 0$ range over non-negative integers, so the left-hand side acts as a pattern: for instance, $A(a+1, 2b+1)$ matches any configuration whose first exponent is positive and whose second exponent is odd. Starting from the blank tape, Antihydra reaches $A(0,4)$ after 11 steps and subsequently follows these macro-transitions. The configuration containing the state $\mathrm{F}$ corresponds to the halting case, as the machine then reaches the undefined $\mathrm{F}0$ transition. Therefore, Antihydra halts if and only if its trajectory from $A(0,4)$ reaches a configuration of the form $A(0,2b+1)$. Whether this ever occurs remains unknown.


\section{Method}

\begin{figure*}[t]
\centering
\includegraphics[width=\textwidth]{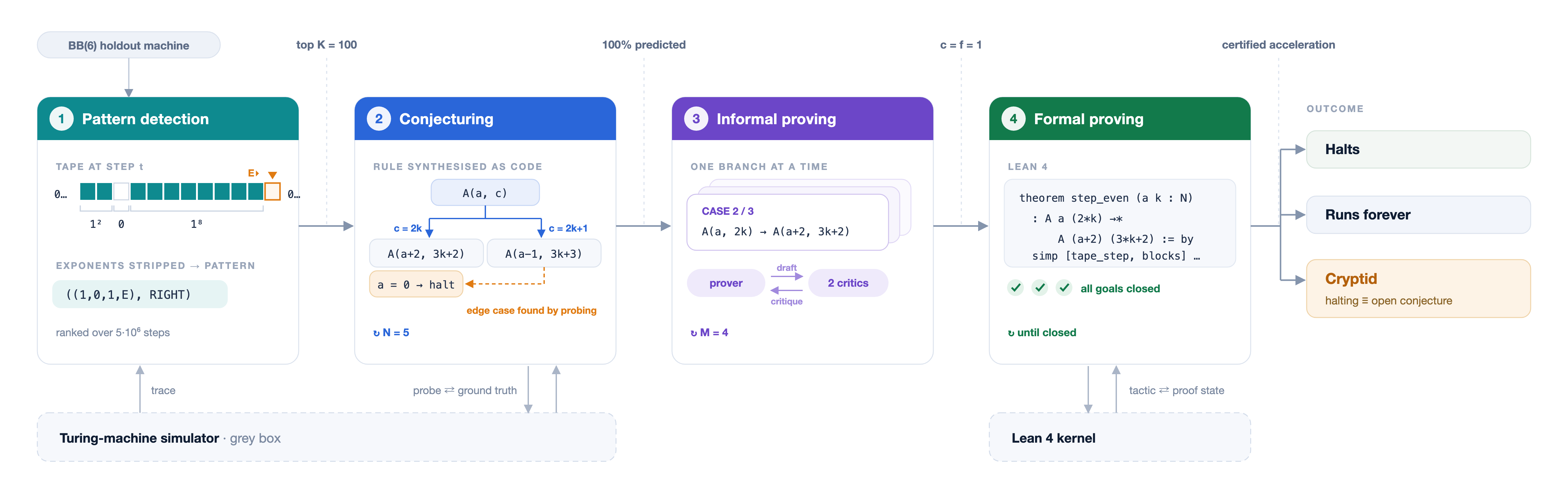}
\caption{Overview of the automated discovery pipeline, detailing the progression from empirical pattern detection and LLM-guided rule conjecturing to informal proving and formal verification in Lean 4.}
\label{fig:pipeline_overview}
\end{figure*}

Our automated discovery pipeline, illustrated in Figure~\ref{fig:pipeline_overview}, operates in four sequential stages, from empirical observation to formal mathematical verification. First, the Pattern Detection phase simulates the Turing machine to isolate frequent abstract tape configurations. Second, the Automated Conjecturing phase employs an LLM to conjecture transition rules over each pattern, which are iteratively refined through targeted simulations. Third, the Informal Proving phase attempts to produce structured mathematical proofs for these conjectured rules. Finally, the Automated Formal Theorem Proving stage translates these informal proofs into machine-checked Lean 4 theorems, guaranteeing the correctness of the discovered long-term dynamics. We present more details on each stage below.


\subsection{Pattern Detection}
\label{subsec:pattern_detection}

The first stage of our discovery pipeline aims to empirically identify repeating patterns in the execution trace of a given Turing machine. To achieve this, the machine is simulated for a fixed number of $S$ steps using a simulator adapted from the \texttt{busy-beaver} repository \citep{ligocki_busy_beaver}. At each simulation step, we record the exact machine configuration: the raw tape contents, active state, and head position.

To convert these raw tape configurations into tractable mathematical representations, we apply a custom compression algorithm (Appendix~\ref{app:tape-compression}). For instance, a raw tape configuration represented in directed head notation as $0^\infty \, 11011111111 \, \text{E}{>} \, 0^\infty$ (where $\text{E}$ denotes the active state and ${>}$ indicates the head direction) is parsed and compressed into a sequence of symbol blocks with corresponding exponents, yielding $0^\infty \, 1^2 \, 0 \, 1^8 \, \text{E}{>} \, 0^\infty$. This algorithm uses a greedy search to parse the finite portions of the tape. To ensure intuitive results, it relies on a simplicity score—defined as the number of repetitions divided by the block size—to favor shorter, highly repetitive sequences over longer, sparser ones. Additionally, it employs a one-step lookahead heuristic to resolve boundary conflicts, intelligently dropping a repetition from a greedy block if doing so allows for greater compression of the subsequent tape segment. Full details and pseudocode are given in Appendix~\ref{app:tape-compression}.

Once a configuration is compressed, we abstract it into a \textit{family of configurations}, which we refer to simply as a \textit{pattern}. A pattern is defined by the sequence of block symbols, the current machine state, and the direction of the read/write head; crucially, the specific block exponents are stripped away. Building on the previous example, the compressed configuration $0^\infty \, 1^2 \, 0 \, 1^8 \, \text{E}{>} \, 0^\infty$ is abstracted into the general pattern $((1, 0, 1, \text{E}), \text{RIGHT})$. 

Throughout the $S$-step simulation, every recorded configuration is grouped into a dataset keyed by its corresponding abstract pattern. Once the simulation is complete, we rank these patterns from most to least common based on the volume of unique configurations they encompass. This ranking successfully isolates the dominant cyclical behaviors of the Turing machine, allowing us to select the top $K$ most frequent patterns for the subsequent automated conjecturing phase.

\subsection{Automated Conjecturing of Transition Rules}
\label{subsec:conjecturing_initial}

Following the identification of frequent patterns, the pipeline attempts to deduce the underlying mathematical rules governing the transitions between instances of the same pattern. For the top $K$ most frequent patterns, we extract a dataset of observed transitions from the initial $S$-step simulation. Each transition consists of a pair of successive configurations belonging to the same pattern, mapping the block exponents of a configuration at the start of a cycle to the block exponents of the next occurrence. For example, using the pattern $((1, 0, 1, \text{E}), \text{RIGHT})$ from the ``Antihydra'' machine, a recorded transition might map the configuration used in our previous example, $0^\infty \, 1^2 \, 0 \, 1^8 \, \text{E}{>} \, 0^\infty$, to its subsequent occurrence $0^\infty \, 1^4 \, 0 \, 1^{14} \, \text{E}{>} \, 0^\infty$. In terms of the block exponents, this corresponds to the mapping $(2, 1, 8) \rightarrow (4, 1, 14)$.

We prompt a Large Language Model (LLM) with this empirical dataset of transitions. The LLM is tasked with writing an executable program (e.g., a Python function) that takes the initial sequence of block exponents as input and deterministically outputs the subsequent sequence of exponents. 

By framing the conjecture as code, we gain a critical advantage: evaluability. We can easily and automatically test whether the synthesized function accurately predicts the next set of exponents by executing it over the entire dataset of recorded transitions and measuring the proportion of outcomes it correctly resolves. Given a pattern parameterized by three block exponents, $(x, y, z)$, the LLM is expected to synthesize a function that encapsulates the transition dynamics. For Antihydra, the middle exponent always equals $1$, and the exponent triple $(a, 1, b)$ corresponds to the configuration $A(a, b)$ of Section~\ref{subsec:background_cryptids}. An example of an initial synthesized function for the Antihydra machine is illustrated below:

\begin{minted}[breaklines,fontsize=\footnotesize]{python}
def transition_rule(x: int, y: int, z: int) -> tuple[int, ...] | None:
    # Case: z is even (A(a, 2b) -> A(a+2, 3b+2))
    if z % 2 == 0:
        return (x + 2, 1, 3 * (z // 2) + 2)
    # Case: z is odd (A(a, 2b+1) -> A(a-1, 3b+3))
    else:
        return (x - 1, 1, 3 * (z // 2) + 3)
\end{minted}

Its two branches correspond to the first and last macro-transitions of Section~\ref{subsec:background_cryptids}, the halting rule $A(0, 2b+1)$ is missing. This synthesized function acts as an initial empirical conjecture for the machine's behavior. However, because this function is strictly derived from the distribution of configurations observed during the initial simulation, it is often incomplete. Specifically, it fails to account for rare edge cases that the machine has not yet naturally encountered. For instance, if the machine never reaches a state where $z$ is odd and $x = 0$ during the initial simulation (and, in practice, this may not happen even after billions of steps), the empirical dataset will lack this scenario. Consequently, the initial function above would erroneously output a negative exponent $(-1)$ instead of correctly identifying the halting condition, necessitating a refinement phase.

\subsection{Interactive Refinement of Conjectures}
\label{subsec:conjecturing_refinement}

Simply extending the initial simulation offers diminishing returns: as the tape blocks grow, observing a single additional Collatz-like transition can require increasingly many underlying Turing machine steps. A much longer execution trace may therefore provide only a few additional examples of the transitions of interest. Moreover, some cases required for a general transition rule may never occur along the blank-tape trajectory.

To overcome these limitations, we equip the LLM with an interactive simulation tool that allows it to test conjectures directly on configurations of its own choosing. When the LLM identifies a potential edge case—or simply requires more data to verify an uncertain hypothesis—it can proactively query this tool by providing a specific set of synthetic exponents. The tool then constructs the precise hypothetical tape configuration corresponding to that pattern instance. For example, if the LLM queries the tool with exponents $x=0, y=1, z=9$, the tool initializes the Turing machine at the custom state $0^\infty \, 1^0 \, 0 \, 1^9 \, \text{E}{>} \, 0^\infty$ and simulates the forward execution until the machine either transitions back to the target pattern, halts, or exceeds a predefined step limit.

Crucially, the model can test configurations without knowing whether they are reachable from the blank tape, probing cases absent from the observed trajectory and discovering missing branches of the transition logic. This capability proved essential to effective conjecture refinement in our pipeline. Continuing our previous example, the agent can use the interactive tool to simulate the exact scenario where $z$ is odd and $x=0$. Observing that the simulation halts rather than yielding invalid negative exponents, the agent successfully identifies the missing behavior and refines its programmatic conjecture to explicitly include the necessary halting condition.

This iterative process of hypothesizing, targeted simulation, and code revision is executed for up to $N$ iterations. Through this loop, the LLM progressively transforms an incomplete empirical guess into a comprehensive transition rule. 

At the conclusion of the $N$ iterations, the agent translates the finalized programmatic rule into a clean, piecewise mathematical function that maps input block exponents to output block exponents. Where possible, the agent also algebraically accelerates the function to encompass multiple transition steps at once. For Cryptid machines, these functions frequently exhibit Collatz-like behavior. For instance, the final mathematical conjecture for the aforementioned pattern is formalized as:
\begin{enumerate}
    \item $A(0, 2k + 1) \rightarrow \mathtt{halt}$ for $k \ge 0$.
    \item $A(a, 2k) \rightarrow A(a + 2, 3k + 2)$ for $a \ge 0, k \ge 1$.
    \item $A(a, 2k + 1) \rightarrow A(a - 1, 3k + 3)$ for $a \ge 1, k \ge 0$.
\end{enumerate}

\subsection{Informal Proving}
\label{subsec:informal_proving}

Before proceeding to rigorous mathematical verification, we apply a strict empirical filter. We evaluate the synthesized Python function against all recorded transitions—both from the initial simulation and the artificially generated data from the interactive exploration tool. Only conjectures that achieve a perfect empirical verification, correctly predicting 100\% of the transitions, are advanced to the informal proving phase. 

Because the conjectured functions are decomposed into distinct cases (e.g., conditions based on exponent parity or specific boundaries), the system isolates each rule and attempts to prove it individually by tracking the abstract machine tape over multiple steps.

To generate informal mathematical proofs without human intervention, we use the same LLM in an iterative generator-evaluator loop. At each iteration, the model is prompted under three distinct roles:
\begin{enumerate}[nosep]
    \item \textbf{Proof Generator:} Drafts an informal, step-by-step mathematical proof demonstrating that the left-hand configuration of a rule deterministically evolves into the right-hand configuration.
    \item \textbf{Correctness Evaluator:} Critiques the mathematical validity of the generated proof. It assigns a score $c \in \{0, 0.5, 1\}$ based on the following rubric:
    \begin{itemize}[nosep]
        \item $1.0$: The proof is completely correct; all abstract Turing machine steps are executed properly and clearly demonstrated.
        \item $0.5$: The proof is generally correct but contains minor calculation errors, or omits intermediate steps.
        \item $0.0$: The proof contains fatal logic errors, hallucinates transitions, or features severe omissions.
    \end{itemize}
    \item \textbf{Format Evaluator:} Assesses the clarity, structure, and readability of the proof. It assigns a score $f \in \{0, 0.5, 1\}$ based on the following rubric:
    \begin{itemize}[nosep]
        \item $1.0$: The proof is perfectly structured, with all steps clearly described according to established mathematical guidelines.
        \item $0.5$: The proof is generally understandable, but certain sections lack sufficient detail or explanation.
        \item $0.0$: The proof is rushed, skimped, or poorly articulated.
    \end{itemize}
\end{enumerate}

If a generated proof does not receive the maximum scores $c=1$ and $f=1$, the evaluators' textual feedback and identified flaws are returned to the model under the Proof Generator role. The model then revises and resubmits the proof. This iterative critique-and-refine cycle continues for up to $M$ iterations, terminating early if both evaluation criteria are satisfied.

Once a proof receives maximum scores from both evaluators, it is accepted as an evaluator-approved informal proof. The process is repeated for each case-specific rule, and the accepted proofs are passed to the final formalization stage, where their correctness is verified in Lean 4.

\subsection{Formal Proving}
\label{subsec:formal_proving}
The informal proving phase establishes strong confidence in the generated conjectures. However, to formally guarantee that the conjectured function captures the Turing machine's long-term behavior, the final stage of our pipeline verifies the results in Lean 4. For this, we build on the BusyLean repository \citep{yuen_busylean}, adapting and extending its definitions and lemmas for reasoning about Turing machines in Lean, and provide them in-context to the agent. This allows the agent to reason about Collatz-like transition rules over abstract families of tape configurations (patterns). We also supply the agent with the informal proofs generated during the previous phase, along with the initial auto-formalized Lean statements about the various transition rules, as additional context. Together, these proofs and formal statements act as a scaffold, guiding the agent as it formally defines and verifies the Turing machine's long-term behavior.

\noindent\textbf{1. Auto-Formalization and Initialization:} The process begins by providing the agent with three sets of formalized statements: an equivalence statement in Lean between the Turing machine and the previously derived set of rules, a system closure statement, and the target halting or non-halting theorems. These are given as initial context to a Gemini-powered multi-persona agent tailored for Lean theorem proving, alongside the candidate machine's definitions and the informally verified closure proofs.

Our approach is fundamentally different from existing auto-formalization approaches explored in Lean. Existing auto-formalization approaches treat the informal math text as ground truth and attempt a one-to-one translation from English into a formal language. However, informal math often omits edge cases, ignores variable bounds, or relies on implicit domain knowledge. Lean's strict type theory checker does not tolerate such handwaving, and so a strict translation of slightly inaccurate informal text inevitably yields an unprovable formal theorem.

To overcome this, our agentic prover treats the initially seeded Lean statements not as rigid specifications, but as guidance. The agent is allowed to modify and refine the initially seeded Lean statements as it searches for a proof. Specifically, it can adjust rules to add missing bounds to parameters or introduce necessary new rules that were omitted from the initial system closure statement. This capability mirrors the workflow of an expert mathematician: attempting a proof, identifying a missing boundary condition, and subsequently refining the initial theorem statement so the proof succeeds. Furthermore, if the agent determines the existing rules are insufficiently comprehensive to define the system closure, it may introduce and prove additional transition rules, and integrate them into the closure statement, as long as these new additions do not trivialize the theorem by introducing false hypotheses.

\noindent\textbf{2. The Multi-Persona Architecture:} To translate, plan, prove, and verify the formal math statements, the system orchestrates four specialized persona roles:

\noindent\textbf{Problem Setter:} Acts as an auto-formalizer, turning the seeded informal proofs and rule statements into Lean theorem statements. Since the initial conjecture and informal proofs may be wrong on edge cases (e.g., a missing bound or an omitted boundary rule), when the Writer or the Critic reports an unprovable statement, the Problem Setter may modify existing rules or add new ones, using simulation tools to test candidate rules against the machine's actual execution. Only the rules may change: the Lean definition of the Turing machine and the start configuration are fixed. The start configuration is the first one reached from the blank tape, which we check by direct simulation. The final system is valid if (i) each rule is proven against the fixed machine, and (ii) the system is closed, i.e., the start configuration and every rule's output either halt or match some rule's input. Any system passing these checks is a correct acceleration of the machine, however much it departs from the initial conjecture.

\noindent\textbf{Planner:} Decomposes the overall task of proving the various theorem statements into smaller lemmas, identifies gaps in existing proofs, and coordinates the workflow between the Writer and the Critic.

\noindent\textbf{Writer:} Constructs the formal proofs for the various theorem statements. It uses the informal proofs as guidance, translating them into Lean.

\noindent\textbf{Critic:} Acts as a verifier, ensuring that the proofs generated by the Writer are valid. The Critic evaluates feedback generated by the Writer during its proof attempts. This is a crucial step in the agent orchestration, as the Writer might be tackling an unprovable statement and passes this feedback to the Problem Setter. The Critic is responsible for granting final approval. It audits the Problem Setter's formalizations for accuracy and inspects the Writer's proofs to ensure that no unproven steps remain, no illegal axioms are used, and no reward hacks are exploited.

\noindent\textbf{3. Orchestration and Error Correction:} These personas operate over a shared workspace through a round-robin schedule, taking sequential turns to coordinate. Crucially, the auto-formalized statements are not static but are part of a bi-directional feedback loop. If the \textbf{Critic} or \textbf{Writer} identifies a flawed or unprovable rule, or if the Lean compiler throws an error, this feedback flows back to the \textbf{Problem Setter}. To diagnose and correct these errors, the agent is equipped with simulation tools to execute the Turing machine from a given pattern and empirically test the rules before the Problem Setter modifies the formal statements.

\noindent\textbf{4. Termination Criteria:} This iterative refinement loop continues until a final state is reached. The agent terminates successfully once the \textbf{Critic} confirms that all required Lean theorems compile without remaining unproven goals (\texttt{sorry}) or unauthorized axioms. Alternatively, it terminates unsuccessfully if a predefined time limit ($72$ hours) is reached.

\section{Experiments and Results}
\subsection{Implementation}
\label{subsec:implementation}

We implement our pipeline using Gemini 3.1 Pro across all agentic components, integrating custom Python simulation tools with the Lean 4 interactive theorem prover.

\paragraph{Pattern Detection and Conjecturing.}
For initial pattern detection, each Turing machine is simulated for $S = 5 \times 10^6$ steps, from which we extract the top $K = 100$ most frequent abstract patterns. Automated conjecturing and programmatic transition rule synthesis are powered by Gemini 3.1 Pro. During interactive refinement, the targeted simulation tool enforces a limit of $128,000$ steps per synthetic transition, and the agentic refinement loop executes for up to $N = 5$ turns.

\paragraph{Informal and Formal Verification.}
Synthesized rules must achieve a $100\%$ prediction accuracy across all empirical and synthetic transition data to pass the initial evaluation filter. The informal proving stage uses Gemini 3.1 Pro for all three roles in the generator-evaluator loop, which is capped at $M=4$ iterations per rule. Finally, formal theorem proving is driven by an autonomous Gemini 3.1 Pro agent interacting directly with the Lean 4.28.1 environment.

\subsection{Experiments}
\label{subsec:experiments}

To evaluate our pipeline, we first executed the Pattern Detection and Automated Conjecturing stages across three distinct sets of Turing machines, restricting the system to a single conjecturing attempt per pattern. We then deployed the system across the active frontier of unclassified machines: 1,094 $\mathrm{BB}(6)$ holdouts compiled by mxdys, the 65 remaining holdouts in $\mathrm{BB}(2,5)$, and the 4 remaining holdouts in $\mathrm{BB}(3,3)$.

This initial broad search identified 52 promising machines (50 of which belonged to the $\mathrm{BB}(6)$ domain), each yielding two or more conjectures passing the strict 100\% empirical accuracy threshold. The complete list is provided in Appendix~\ref{app:candidates}. To increase our chances of generating conjectures that could be formally proven, we isolated these 52 machines and re-ran the conjecturing pipeline with higher-resolution settings, raising the allowance to 4 conjecturing attempts per pattern.

All conjectures from these high-resolution runs that maintained 100\% empirical accuracy were advanced to the informal proving phase. Conjectures that successfully satisfied both the correctness and format evaluations were then submitted to the autonomous agent for formal theorem proving in Lean.

\subsection{Results}
\label{subsec:results}

Following this rigorous end-to-end verification, our system established formal proofs governing the behavior of several machines. We manually validated all final Lean statements, including the machine definitions, start configurations, and closure theorems. All Lean proofs are available in our \href{https://github.com/google-deepmind/cryptids}{public GitHub repository}. Table \ref{tab:merged-cryptid-results} details these findings across the $\mathrm{BB}(6)$ and $\mathrm{BB}(2,5)$ domains. We categorize our results by \textbf{Discovery type} to highlight the specific contribution relative to existing human knowledge:
\begin{itemize}
    \item \textbf{New Discoveries:} Novel accelerations for 11 machines for which no comparable analysis had previously been documented by the community.
    \item \textbf{Rediscoveries:} Higher-level or alternative accelerations for 15 machines whose behavior had previously been described through related conjectured accelerations by community contributors, including on Discord, but had not been formally verified.
\end{itemize}

The complete systems of transition rules derived by our pipeline for these machines are provided in \cref{app:discovered-accelerations}.


\section{Related Work}
Recent advancements in neural theorem proving have demonstrated remarkable capabilities in both formal and informal mathematical reasoning. Over the past two years, state-of-the-art systems have achieved historic milestones, including reaching silver and gold medal levels at the International Mathematical Olympiad and attaining near-perfect scores on the Putnam exam \citep{hubert2026olympiad, shao2025deepseekmath, liu2026numina}. Pushing beyond these standardized benchmarks, frontier models and systems have successfully resolved open mathematical questions \citep{feng2026towards,alphaproof_nexus,alon2026remarks}. While these systems excel at navigating complex search spaces to find proofs, they were not designed to autonomously generate novel mathematical conjectures or formulate new theories. Beyond problem solving, frameworks such as the AI Co-Mathematician \citep{zheng2026ai} leverage AI agents for open-ended research through ideation, literature search, computational exploration, theorem proving, and theory building. However, they function primarily as collaborative assistants that rely on human intuition to define the mathematical objects of interest and guide the direction of exploration. Taking a fully autonomous approach, evolutionary systems like AlphaEvolve \citep{novikov2025alphaevolve} explore open-ended spaces by mutating code to discover novel algorithmic heuristics. However, their strict reliance on a climbable reward signal fundamentally restricts their application to general mathematics, where such smooth, quantifiable optimization gradients rarely exist.

Recently, several AI systems have emerged to automate the scientific discovery process, spanning from hypothesis generation \citep{gottweis2026accelerating,zheng2026ai} to experimental analysis \citep{zheng2025automation}. Notably, Sakana AI \citep{lu2026towards} introduced an end-to-end automated scientist for machine learning that autonomously conducts research—from ideation and coding to manuscript writing and peer review—ultimately producing a workshop-accepted paper. While such frameworks are theoretically generalizable, empirical domains like machine learning are inherently easier to evaluate than mathematics, because in the latter establishing a conjecture requires a mathematical proof, although experiments can help test and refine it. To address this challenge, newer frameworks divide mathematical discovery into distinct phases: conjecture generation and proof writing. As an example, ConjectureBench~\citep{conjecturebench2026} isolates the conjecturing phase, producing statements that automated theorem provers can then attempt to verify. Building on this, AI systems like LEAP~\citep{kung2026leap} can  autonomously attempt to prove theorems by pairing an ideation agent that drafts informal proofs with a formalization agent that translates them into verified Lean 4 code, marking significant progress toward end-to-end automated discovery in mathematics.

Applying these powerful AI-driven methodologies provides a new approach to classical challenges in computability theory, most notably the Busy Beaver problem originally introduced by Radó \citep{rado1962non}. Historically, evaluating these theoretical limits has relied on exhaustive manual heuristics to filter out non-halting Turing machines, leaving behind complex "holdouts" whose behaviors are intimately tied to deep mathematical conjectures \citep{michel2009busy, aaronson2020busy}. The implications of these limits reveal the uncomputable, open-ended nature of algorithmic generation \citep{chaitin2013proving}. Despite this immense computational complexity, the recent monumental proof resolving the fifth Busy Beaver value was achieved entirely by a collaborative community effort without the use of artificial intelligence \citepalias{blanchard2026determination}. More recently, in other models of computation, AI is already making direct mathematical contributions; most notably, the large language model Claude Opus 4.6 was recently used to successfully generate proofs of non-halting for hundreds of complex FRACTRAN holdout machines, entirely resolving the remaining $\mathrm{BB}_f(21)$ cases and proving all but the 'Fenrir-family' Cryptids in the $\mathrm{BB}_f(22)$ domain (\href{https://wiki.bbchallenge.org/wiki/Fractran}{Wiki}).

\vspace{-0.5cm}
\section{Conclusion}
In this work, we explored the Busy Beaver frontier as an open-ended testbed for automated mathematical discovery and formal verification. We introduced a pipeline that combines empirical pattern detection, LLM-guided conjecturing, and informal proving through a generator-evaluator loop to derive Collatz-like accelerations for Turing machines, expressed as systems of transition rules between parametrized tape configurations. These transition rules are then formally verified in Lean 4, establishing the correctness of each acceleration.

The empirical success of our pipeline was demonstrated by the discovery and formal verification of accelerations within the $\mathrm{BB}(6)$ and $\mathrm{BB}(2,5)$ domains. Our system successfully derived and formally verified novel Collatz-like accelerations for eleven machines in $\mathrm{BB}(6)$. Subject to community acceptance, we classify these machines as Cryptids in the sense of \cref{def:cryptid}. These results represent a step toward systematically classifying the remaining holdouts in the Busy Beaver search.

By design, the Busy Beaver domain is fundamentally open-ended due to its uncomputability, making it a compelling testbed for open-ended AI research \citep{hughes2024open}. However, we acknowledge that our current approach is not entirely open-ended. Specifically, our system currently accelerates machines exhibiting Collatz-like functions—structures already studied by mathematicians and whose long-term behavior can be exceptionally difficult to resolve. A more ambitious version of this work would relax these assumptions, expanding beyond Collatz-like dynamics to other unsolved mathematical problems. The ultimate goal would be a fully autonomous system capable of discovering entirely novel classes of open problems and mapping Turing machines to them.

While the Busy Beaver challenge acts as a generator of mathematical questions—given that important open problems can be encoded as Turing machines whose halting status must ultimately be resolved—the broader mathematical significance of the questions arising from its exploration is not guaranteed. For instance, although progress on specific Cryptids could theoretically yield insights into the Collatz conjecture, such outcomes remain speculative and cannot be guaranteed. Regardless of these domain-specific caveats, our architecture serves as a blueprint for broader AI-driven mathematical discovery. By equipping large language models with interactive computational tools, generator-evaluator critique loops, and iterative formal proving environments, we add to the growing body of work on autonomous systems capable of tackling increasingly profound challenges in theoretical computer science.

\vspace{-0.3cm}
\section{Acknowledgments}
We thank Mateja Jamnik, Wenda Li, and Albert Q. Jiang for advising the main author on his PhD thesis. We thank Paul Lezeau for his help with formalizing the directed head notation in Lean and for developing the Lean formalization of Turing machines within the Formal Conjectures project. We would also like to thank our colleagues at Google DeepMind for sharing invaluable feedback on this work.

\newpage
\bibliography{iclr2026_conference}

\newpage
\appendix
\onecolumn

\section{Accelerations Discovered by Our Pipeline}
\label[appendix]{app:discovered-accelerations}
This section presents the systems of transition rules discovered and formally verified by our pipeline for eleven previously unanalyzed machines in $\mathrm{BB}(6)$. The corresponding Lean proofs are available in our \href{https://github.com/google-deepmind/cryptids}{public GitHub repository}.

\subsection{New Discovery: \texttt{1RB1LD\_\allowbreak 1RC0RA\_\allowbreak 1LA1RE\_\allowbreak 0LA0LB\_\allowbreak 0RB1RF\_\allowbreak -{}-{}-1RD}}
\texttt{1RB1LD\_\allowbreak 1RC0RA\_\allowbreak 1LA1RE\_\allowbreak 0LA0LB\_\allowbreak 0RB1RF\_\allowbreak -{}-{}-1RD} is a holdout in $\mathrm{BB}(6)$ for which no acceleration had previously been documented by the community. Our pipeline discovered and formally verified the following Collatz-like acceleration for this machine:
\begin{minted}[breaklines, fontsize=\footnotesize]{text}
Consider:
A(a, b, c, d) = 0^inf 1^a 10^b A> 01^c 1^d 0^inf

A(a, b, c, d) -> A(a, b+1, c-1, d) if c > 0
A(a, b, 0, 0) -> A(a+1, 0, 0, 2b+2)
A(a, b, 0, d) -> A(a-3, b+2, 0, d-1) if d > 0 and a >= 3
A(2, b, 0, d) -> A(0, 0, b+1, d-1) if d > 0
A(1, b, 0, d) -> A(1, 1, b+1, d-1) if d > 0 and b > 0
A(1, 0, 0, d) -> A(0, 0, 0, d+2) if d > 0
A(0, 0, 0, 1) -> A(0, 0, 2, 5)
A(0, b, 0, 1) -> A(0, 0, 2, 2b+1) if b > 0
A(0, b, 0, d) -> A(3d-4, 1, 2, 2b-2d+3) if d >= 2 and b >= d-1
A(0, b, 0, b+3) -> A(3b+2, 0, 0, 6)
A(0, b, 0, b+4) -> A(3b+4, 0, 0, 4)
A(0, b, 0, d) -> A(3b+3, 2, 0, d-b-5) if d >= b+5
A(0, b, 0, b+2) -> Halt

Start: A(0, 0, 0, 0), already present on the initial blank tape
\end{minted}

\subsection{New Discovery: \texttt{1RB1RD\_\allowbreak 0LC1RA\_\allowbreak 0RA1LC\_\allowbreak 1RA1LE\_\allowbreak 1RF0LD\_\allowbreak -{}-{}-0RE}}
\texttt{1RB1RD\_\allowbreak 0LC1RA\_\allowbreak 0RA1LC\_\allowbreak 1RA1LE\_\allowbreak 1RF0LD\_\allowbreak -{}-{}-0RE} is a holdout in $\mathrm{BB}(6)$ for which no acceleration had previously been documented by the community. Our pipeline discovered and formally verified the following Collatz-like acceleration for this machine:

\begin{minted}[breaklines, fontsize=\footnotesize]{text}
Consider:
P_1(c) = 0^inf C> 0 1^c 0^inf
P_2(a, c) = 0^inf 10^a 1 C> 0 1^c 0^inf
D(a, b, w) = 0^inf 10^a 1^b D> 1^{w+1} 0^inf
where c >= 1

P_1(1) -> P_1(3)
P_1(w+2) -> D(0, 1, w)
P_2(a, 1) -> P_2(a, 3)
P_2(a, w+2) -> D(a+1, 1, w)

D(0, 2k+1, 0) -> P_1(2k+4)
D(0, 2k+1, 1) -> P_1(2k+6)
D(0, 2k+1, w+2) -> D(0, 2k+4, w)

D(0, 2k+2, 0) -> Halt
D(0, 2k+2, 1) -> P_2(k, 3)
D(0, 2k+2, 2) -> P_2(k, 5)
D(0, 2k+2, 3) -> P_2(k, 7)
D(0, 2k+2, w+4) -> D(k+1, 5, w)

D(1, 2k+1, 0) -> P_1(2k+5)
D(1, 2k+1, 1) -> P_1(2k+7)
D(a+2, 2k+1, 0) -> P_2(a, 2k+5)
D(a+2, 2k+1, 1) -> P_2(a, 2k+7)
D(a+1, 2k+1, w+2) -> D(a, 2k+5, w)

Start: P_1(1), reached after 3 raw TM steps
\end{minted}

This initial acceleration can be simplified by composing transitions through $P_1$ and $P_2$, yielding a system expressed in terms of the configuration $D$:
\begin{minted}[breaklines, fontsize=\footnotesize]{text}
Consider:
D(a, b, w) = 0^inf 10^a 1^b D> 1^{w+1} 0^inf where b >= 1, and b is odd if a >= 1

D(0, 2k+1, w) -> D(0, 1, 2k+2w+2) if w <= 1
D(0, 2k+1, w+2) -> D(0, 2k+4, w)
D(0, 2k+2, 0) -> Halt
D(0, 2k+2, w) -> D(k+1, 1, 2w-1) if 1 <= w <= 3
D(0, 2k+2, w+4) -> D(k+1, 5, w)
D(a+1, 2k+1, w) -> D(a, 1, 2k+2w+3) if w <= 1
D(a+1, 2k+1, w+2) -> D(a, 2k+5, w)

Start: D(0, 1, 1), reached after 13 raw TM steps
\end{minted}

\subsection{New Discovery: \texttt{1RB1RB\_\allowbreak 1LC0RE\_\allowbreak -{}-{}-1LD\_\allowbreak 0RA1LB\_\allowbreak 1RA1LF\_\allowbreak 0LE1LE}}
\texttt{1RB1RB\_\allowbreak 1LC0RE\_\allowbreak -{}-{}-1LD\_\allowbreak 0RA1LB\_\allowbreak 1RA1LF\_\allowbreak 0LE1LE} is a holdout in $\mathrm{BB}(6)$ for which no acceleration had previously been documented by the community. Our pipeline discovered and formally verified the following Collatz-like acceleration for this machine:
\begin{minted}[breaklines, fontsize=\footnotesize]{text}
Consider:
C(a, c, f) = 0^inf 1^a 0 1^c 0 1 <D 1^f 0^inf where a, c are even, a >= 2, c >= 2, f >= 1

C(2k, 2m, f) -> C(2, 2m-2k+2, f+6k-6) if 2 <= k < m, f >= 1
C(2k, 2m, f) -> C(2k-2m+2, 2, f+6m-6) if 2 <= m < k, f >= 1
C(2k, 2k, f) -> C(2, 2, f+6k-6) if k >= 2, f >= 1
C(2k, 2, 2j)   -> C(2k-4, 2j+8, 2) if k >= 3, j >= 1
C(2k, 2, 2j+1) -> C(2k-4, 2j+8, 5) if k >= 3, j >= 0
C(4, 2, 2j)    -> C(2j+4, 6, 5) if j >= 1
C(4, 2, 2j+1)  -> C(2j+8, 2, 2) if j >= 0
C(2, 2m, 2j)   -> C(2m-6, 2j+10, 5) if m >= 4, j >= 1
C(2, 2m, 2j+1) -> C(2m-6, 2j+12, 2) if m >= 4, j >= 0
C(2, 6, 2j)    -> C(2j+10, 2, 2) if j >= 1
C(2, 6, 2j+1)  -> C(2j+8, 6, 5) if j >= 0
C(2, 4, f) -> Halt if f >= 1
C(2, 2, f) -> Halt if f >= 1

Start: C(6, 2, 2), reached after 76 raw TM steps
\end{minted}

\subsection{New Discovery: \texttt{1RB0LF\_\allowbreak 1LC1RA\_\allowbreak 0RE0RD\_\allowbreak -{}-{}-1LE\_\allowbreak 1LF1RC\_\allowbreak 1LC1LA}}
\texttt{1RB0LF\_\allowbreak 1LC1RA\_\allowbreak 0RE0RD\_\allowbreak -{}-{}-1LE\_\allowbreak 1LF1RC\_\allowbreak 1LC1LA} is a holdout in $\mathrm{BB}(6)$ for which no acceleration had previously been documented by the community. Our pipeline discovered and formally verified the following Collatz-like acceleration for this machine:
\begin{minted}[breaklines, fontsize=\footnotesize]{text}
Consider:
C(a, b) = 0^inf E> 10^a 1^b 0^inf

C(a, b) -> C(2a+2, b-2a-2) if b > 2a+2
C(a, 1) -> C(2, 2a+2)
C(a, 2)  -> Halt
C(a, 2k) -> C((6a+2k+2)/4, (6a+2k+2)%4+1) if a >= 1 and 2 <= k <= a+1
C(a, 2k+1) -> C((6a+2k+6)/4, (6a+2k+6)%4+1) if a >= 1 and 1 <= k <= a

Start: C(0,3), reached after 7 raw TM steps
\end{minted}

\subsection{New Discovery: \texttt{1RB1RD\_\allowbreak 0LC1RA\_\allowbreak 1RA1LC\_\allowbreak 0RB1LE\_\allowbreak 1RF0LD\_\allowbreak -{}-{}-0RE}}
\texttt{1RB1RD\_\allowbreak 0LC1RA\_\allowbreak 1RA1LC\_\allowbreak 0RB1LE\_\allowbreak 1RF0LD\_\allowbreak -{}-{}-0RE} is a holdout in $\mathrm{BB}(6)$ for which no acceleration had previously been documented by the community. Our pipeline discovered and formally verified the following Collatz-like acceleration for this machine:
\begin{minted}[breaklines, fontsize=\footnotesize]{text}
Consider:
P_1(c) = 0^inf C> 0 1^c 0^inf
P_2(a, c) = 0^inf 10^a 1 C> 0 1^c 0^inf
D_1(a, z, w) = 0^inf 10^a 1^z D> 1 1^w 0^inf

P_1(1) -> P_1(4)
P_2(0, 1) -> P_1(5)
P_2(a, 1) -> P_2(a-1, 5) if a >= 1
P_1(c) -> D_1(0, 2, c-2) if c >= 2
P_2(a, c) -> D_1(a, 3, c-2) if c >= 2

D_1(0, 2k+1, w) -> D_1(0, 2k+4, w-2) if w >= 2
D_1(0, 2k+2, w) -> D_1(k+1, 5, w-4) if w >= 4
D_1(a, 2k+1, w) -> D_1(a-1, 2k+5, w-2) if a >= 1 and w >= 2

D_1(0, 2k+1, 0) -> P_1(2k+4)
D_1(0, 2k+1, 1) -> P_1(2k+6)
D_1(0, 2k+2, 0) -> Halt
D_1(0, 2k+2, 1) -> P_2(k, 3)
D_1(0, 2k+2, 2) -> P_2(k, 5)
D_1(0, 2k+2, 3) -> P_2(k, 7)
D_1(1, 2k+1, 0) -> P_1(2k+5)
D_1(1, 2k+1, 1) -> P_1(2k+7)
D_1(a, 2k+1, 0) -> P_2(a-2, 2k+5) if a >= 2
D_1(a, 2k+1, 1) -> P_2(a-2, 2k+7) if a >= 2

Start: P_1(1), reached after 3 raw TM steps
\end{minted}

This initial acceleration can be simplified by composing transitions through $P_1$ and $P_2$, yielding a system expressed in terms of the configuration $D_1$:
\begin{minted}[breaklines, fontsize=\footnotesize]{text}
Consider:
D_1(a, z, w) = 0^inf 10^a 1^z D> 1 1^w 0^inf

D_1(0, 2k+1, w) -> D_1(0, 2k+4, w-2) if w >= 2
D_1(0, 2k+2, w) -> D_1(k+1, 5, w-4) if w >= 4
D_1(a, 2k+1, w) -> D_1(a-1, 2k+5, w-2) if a >= 1 and w >= 2

D_1(0, 2k+1, 0) -> D_1(0, 2, 2k+2)
D_1(0, 2k+1, 1) -> D_1(0, 2, 2k+4)
D_1(0, 2k+2, 0) -> Halt
D_1(0, 2k+2, 1) -> D_1(k, 3, 1)
D_1(0, 2k+2, 2) -> D_1(k, 3, 3)
D_1(0, 2k+2, 3) -> D_1(k, 3, 5)
D_1(1, 2k+1, 0) -> D_1(0, 2, 2k+3)
D_1(1, 2k+1, 1) -> D_1(0, 2, 2k+5)
D_1(a, 2k+1, 0) -> D_1(a-2, 3, 2k+3) if a >= 2
D_1(a, 2k+1, 1) -> D_1(a-2, 3, 2k+5) if a >= 2

Start: D_1(0, 2, 2), reached after 16 raw TM steps
\end{minted}

\subsection{New Discovery: \texttt{1RB0LF\_\allowbreak 1RC1RA\_\allowbreak 0RD0RB\_\allowbreak 1LE-{}-{}-\_\allowbreak 0LA1LE\_\allowbreak 1RD1LA}}
\texttt{1RB0LF\_\allowbreak 1RC1RA\_\allowbreak 0RD0RB\_\allowbreak 1LE-{}-{}-\_\allowbreak 0LA1LE\_\allowbreak 1RD1LA} is a holdout in $\mathrm{BB}(6)$ for which no acceleration had previously been documented by the community. Our pipeline discovered and formally verified the following Collatz-like acceleration for this machine:
\begin{minted}[breaklines, fontsize=\footnotesize]{text}
Consider:
J(u, v, w) = 0^inf 1^5 10^u 1 A> 11 01^v 1^w 01 0^inf
S_E(b, m, k) = 0^inf 1^5 10^b E> 1 1101^m 1^6 11^k 01 0^inf
S_{E,0}(b, m) = 0^inf 1^5 10^b E> 1 1101^m 1101 0^inf
S_{E,1}(b, m, n) = 0^inf 1^5 10^b E> 1 1101^m 111 01^n 01001 0^inf

J(0, v, 0) -> S_{E,0}(v+3, 0)
J(0, v, 1) -> S_{E,1}(v+2, 0, 0)
J(0, v, 2) -> S_E(v+2, 0, 0)
J(0, v, w) -> J(v+3, 0, w-3) if w >= 3
J(1, v, 0) -> S_E(0, 0, v+3)
J(u, v, 0) -> S_{E,1}(u-2, 0, v+2) if u >= 2
J(1, v, 1) -> Halt
J(u, v, 1) -> S_E(u-2, 0, v+2) if u >= 2
J(u, v, w) -> J(u-1, v+2, w-2) if u >= 1 and w >= 2

S_E(0, 0, k) -> J(0, 0, 2k+6)
S_E(0, m, k) -> J(2, 2m+2, 2k+1) if m >= 1
S_E(1, m, k) -> J(0, 2m+3, 2k+1)
S_E(b, m, k) -> S_E(b-2, m+1, k) if b >= 2

S_{E,0}(0, m) -> S_E(2, 0, 2m)
S_{E,0}(1, m) -> S_E(0, 0, 2m+1)
S_{E,0}(b, m) -> S_{E,0}(b-2, m+1) if b >= 2

S_{E,1}(0, m, n) -> S_E(2, 0, 2m+n+2)
S_{E,1}(1, m, n) -> S_E(0, 0, 2m+n+3)
S_{E,1}(b, m, n) -> S_{E,1}(b-2, m+1, n) if b >= 2

Start: J(0, 0, 0), reached after 35 raw TM steps
\end{minted}

This initial acceleration can be simplified by composing transitions through $S_E$, $S_{E,0}$, and $S_{E,1}$, yielding a system expressed in terms of the configuration $J$:
\begin{minted}[breaklines, fontsize=\footnotesize]{text}
Consider:
J(u, v, w) = 0^inf 1^5 10^u 1 A> 11 01^v 1^w 01 0^inf

J(0, v, 0) -> J(0, 0, 2v+12) if v is even
J(0, v, 0) -> J(2, 4, 2v+7) if v is odd
J(0, v, 1) -> J(2, 4, 2v+9) if v is even
J(0, v, 1) -> J(0, 0, 2v+14) if v is odd
J(0, v, 2) -> J(2, v+4, 1) if v is even
J(0, v, 2) -> J(0, v+4, 1) if v is odd
J(0, v, w) -> J(v+3, 0, w-3) if w >= 3

J(1, v, 0) -> J(0, 0, 2v+12)
J(u, v, 0) -> J(2, 4, 2u+2v+5) if u >= 2 and u is even
J(u, v, 0) -> J(0, 0, 2u+2v+10) if u >= 3 and u is odd

J(1, v, 1) -> Halt
J(2, v, 1) -> J(0, 0, 2v+10)
J(u, v, 1) -> J(2, u, 2v+5) if u >= 4 and u is even
J(u, v, 1) -> J(0, u, 2v+5) if u >= 3 and u is odd

J(u, v, w) -> J(u-1, v+2, w-2) if u >= 1 and w >= 2

Start: J(0, 0, 0), reached after 35 raw TM steps
\end{minted}

\subsection{New Discovery: \texttt{1RB0LC\_\allowbreak 0LC1RF\_\allowbreak 1LE1RD\_\allowbreak 0RC1RC\_\allowbreak 0LA1LA\_\allowbreak -{}-{}-1RA}}
\texttt{1RB0LC\_\allowbreak 0LC1RF\_\allowbreak 1LE1RD\_\allowbreak 0RC1RC\_\allowbreak 0LA1LA\_\allowbreak -{}-{}-1RA} is a holdout in $\mathrm{BB}(6)$ for which no acceleration had previously been documented by the community. Our pipeline discovered and formally verified the following Collatz-like acceleration for this machine:
\begin{minted}[breaklines, fontsize=\footnotesize]{text}
Consider:
E(a, k, c) = 0^inf 1^a A> 0 11^k 0 1^c 0^inf
M_1(a, k, z) = 0^inf 1^a 0 11^k 1 0 C> 1^z 0^inf
M_0(n, z) = 0^inf 1^n 1 0 C> 1^z 0^inf

E(a, 0, c) -> M_0(a, c)
E(a, 1, 2m) -> E(a+3, m, 0)
E(a, 1, 1) -> Halt
E(a, 1, 2j+3) -> M_0(a+5, 2j)
E(a, k+2, 0) -> M_0(a+2, 2k+1)
E(a, k+2, c) -> M_1(a+3, k, c) if c >= 1

M_1(a, k, 2j+1) -> E(a+3, k, 2j+1)
M_1(a, k, 2) -> E(a, 1, 2k+2)
M_1(a, k, 2j+4) -> E(a+3, k+1, 2j+2)

M_0(n, 0) -> E(3, 0, n+2)
M_0(2m, 2j+1) -> E(3, m, 2j+1)
M_0(2m+1, 2j+1) -> E(4, m+1, 2j+1)
M_0(2m, 2) -> E(0, 1, 2m+2)
M_0(2m+1, 2) -> E(1, 1, 2m+4)
M_0(2m, 2j+4) -> E(3, m+1, 2j+2)
M_0(2m+1, 2j+4) -> E(4, m+2, 2j+2)

Start: E(3, 0, 2), reached after 23 raw TM steps
\end{minted}

This initial acceleration can be simplified by composing transitions through $M_1$ and $M_0$, yielding a system expressed in terms of the configuration $E$:
\begin{minted}[breaklines, fontsize=\footnotesize]{text}
Consider:
E(a, k, c) = 0^inf 1^a A> 0 11^k 0 1^c 0^inf

E(a, 0, 0) -> E(3, 0, a+2)
E(2m, 0, 2j+1) -> E(3, m, 2j+1)
E(2m+1, 0, 2j+1) -> E(4, m+1, 2j+1)
E(2m, 0, 2) -> E(0, 1, 2m+2)
E(2m+1, 0, 2) -> E(1, 1, 2m+4)
E(2m, 0, 2j+4) -> E(3, m+1, 2j+2)
E(2m+1, 0, 2j+4) -> E(4, m+2, 2j+2)

E(a, 1, 2m) -> E(a+3, m, 0)
E(a, 1, 1) -> Halt
E(a, 1, 3) -> E(3, 0, a+7)
E(2m, 1, 5) -> E(1, 1, 2m+8)
E(2m+1, 1, 5) -> E(0, 1, 2m+8)
E(2m, 1, 2j+7) -> E(4, m+4, 2j+2)
E(2m+1, 1, 2j+7) -> E(3, m+4, 2j+2)

E(2m, k+2, 0) -> E(3, m+1, 2k+1)
E(2m+1, k+2, 0) -> E(4, m+2, 2k+1)
E(a, k+2, 2j+1) -> E(a+6, k, 2j+1)
E(a, k+2, 2) -> E(a+3, 1, 2k+2)
E(a, k+2, 2j+4) -> E(a+6, k+1, 2j+2)

Start: E(3, 0, 2), reached after 23 raw TM steps
\end{minted}

\subsection{New Discovery: \texttt{1RB1LF\_\allowbreak 0RC1RD\_\allowbreak 1LA1LD\_\allowbreak 1LE0RA\_\allowbreak -{}-{}-1LC\_\allowbreak 0LA1LA}}
\texttt{1RB1LF\_\allowbreak 0RC1RD\_\allowbreak 1LA1LD\_\allowbreak 1LE0RA\_\allowbreak -{}-{}-1LC\_\allowbreak 0LA1LA} is a holdout in $\mathrm{BB}(6)$ for which no acceleration had previously been documented by the community. Our pipeline discovered and formally verified the following Collatz-like acceleration for this machine:
\begin{minted}[breaklines, fontsize=\footnotesize]{text}
Consider:
C(a, b) = 0^inf 110 A> 11^a 0 1^b 0^inf with b >= 6
M_3(a, c, d) = 0^inf 110 A> 11^a 01 11^c 0 1^d 0^inf
M_4(a, b, c, d) = 0^inf 11^{a+1} 110 A> 11^b 01 11^c 0 1^d 0^inf

Notice that C(a, 2c+1) = M_3(a, c, 0)

C(0, 2k+6) -> C(k+1, 9)
C(a+1, 2k+6) -> M_4(a, k+1, 4, 0)

M_3(0, 0, d) -> Halt
M_3(1, 0, d) -> Halt
M_3(a+2, 0, d) -> C(a+1, d+9)
M_3(0, c+1, d) -> C(c+1, d+6)
M_3(a+1, c+1, d) -> M_3(a, c, d+6)

M_4(a, 0, c, d) -> C(a+c+1, d+6)
M_4(0, 1, 0, d) -> Halt
M_4(a+1, 1, 0, d) -> C(a+1, d+12)
M_4(a, b+2, 0, d) -> M_3(a, b, d+9)
M_4(a, b+1, c+1, d) -> M_4(a, b, c, d+6)

Start: C(0, 9), reached after 74 raw TM steps
\end{minted}

This initial acceleration can be simplified by composing transitions through $M_3$ and $M_4$, yielding a system expressed in terms of the configuration $C$:
\begin{minted}[breaklines, fontsize=\footnotesize]{text}
Consider:
C(a, b) = 0^inf 110 A> 11^a 0 1^b 0^inf with b >= 6

C(0, 2k+6) -> C(k+1, 9)
C(a+1, 2k+6) -> C(a-k+4, 6k+12) if k <= a+3
C(a+1, 2k+6) -> Halt if k = a+4 or k = a+5
C(a+1, 2k+6) -> C(k-a-5, 6a+39) if k >= a+6

C(a, 2k+1) -> C(k-a, 6a+6) if k > a
C(a, 2k+1) -> Halt if a = k or a = k+1
C(a, 2k+1) -> C(a-k-1, 6k+9) if a >= k+2

Start: C(0, 9), reached after 74 raw TM steps
\end{minted}

\subsection{New Discovery: \texttt{1RB1RC\_\allowbreak 1LC1LC\_\allowbreak 1RE0LD\_\allowbreak 1LB1RF\_\allowbreak -{}-{}-1RA\_\allowbreak 0RD1RD}}
\texttt{1RB1RC\_\allowbreak 1LC1LC\_\allowbreak 1RE0LD\_\allowbreak 1LB1RF\_\allowbreak -{}-{}-1RA\_\allowbreak 0RD1RD} is a holdout in $\mathrm{BB}(6)$ for which no acceleration had previously been documented by the community. Our pipeline discovered and formally verified the following Collatz-like acceleration for this machine:
\begin{minted}[breaklines, fontsize=\footnotesize]{text}
Consider:
C_0(a, b, c) = 0^inf 1^{2a+2} E> 1^{2b+2} 0 1^{2c+2} 0^inf
C_1(a, b, c) = 0^inf 1^{2a+3} E> 1^{2b+2} 0 1^{2c+2} 0^inf

C_0(a, 0, 0) -> Halt
C_0(a, 0, 1) -> C_1(2, 2, a+1)
C_0(a, 0, c+2) -> C_0(1, a+3, c)
C_0(a, b+1, 0) -> C_1(a+1, 0, b)
C_0(a, b+1, c+1) -> C_0(a+3, b, c)

C_1(a, 0, 0) -> Halt
C_1(a, 0, 1) -> C_0(1, 0, a+3)
C_1(a, 0, c+2) -> C_1(2, a+3, c)
C_1(a, b+1, 0) -> C_0(a+2, 0, b)
C_1(a, b+1, c+1) -> C_1(a+3, b, c)

Start: C_1(2, 2, 0), reached after 100 raw TM steps
\end{minted}

This acceleration can be simplified by merging the two configuration families, using the reparameterization $C_0(a,b,c) = C(2a,b,c)$ and $C_1(a,b,c) = C(2a+1,b,c)$:
\begin{minted}[breaklines, fontsize=\footnotesize]{text}
Consider:
C(a, b, c) = 0^inf 1^{a+2} E> 1^{2b+2} 0 1^{2c+2} 0^inf

C(a, 0, 0) -> Halt
C(2k, 0, 1) -> C(5, 2, k+1)
C(2k+1, 0, 1) -> C(2, 0, k+3)
C(2k, 0, c+2) -> C(2, k+3, c)
C(2k+1, 0, c+2) -> C(5, k+3, c)

C(a, b+1, 0) -> C(a+3, 0, b)
C(a, b+1, c+1) -> C(a+6, b, c)

Start: C(5, 2, 0), reached after 100 raw TM steps
\end{minted}

\subsection{New Discovery: \texttt{1RB1LE\_\allowbreak 1RC1RA\_\allowbreak 0LD1RB\_\allowbreak 1LE1LD\_\allowbreak 1RF0LA\_\allowbreak -{}-{}-0RE} }
\texttt{1RB1LE\_\allowbreak 1RC1RA\_\allowbreak 0LD1RB\_\allowbreak 1LE1LD\_\allowbreak 1RF0LA\_\allowbreak -{}-{}-0RE}  is a holdout in $\mathrm{BB}(6)$ for which no acceleration had previously been documented by the community. Our pipeline discovered and formally verified the following Collatz-like acceleration for this machine:
\begin{minted}[breaklines, fontsize=\footnotesize]{text}
Consider:
D(a, b, c) = 0^inf 10^b 1^{2a+2} D> 1^{c+1} 0^inf
E(a, b, c) = 0^inf 10^b 1^{2a+2} E> 1^{c+1} 0^inf

D(a+1, b, c) -> D(a, b, c+2)
D(0, 0, 0) -> D(2, 0, 0)
D(0, 0, c+1) -> E(1, 0, c+1)
D(0, 1, c) -> D(c, 2, 0) if c <= 2
D(0, 1, c+3) -> E(1, 2, c+1)
D(0, b+2, c) -> E(1, b, c+2)

E(a+1, b, 0) -> E(a, b, 1)
E(0, 0, 0) -> D(0, 0, 1)
E(0, b+1, 0) -> D(1, b, 0)
E(a, 0, 1) -> D(a+1, 0, 1)
E(a, 0, 2) -> D(a+2, 0, 1)
E(a, 0, 3) -> Halt
E(a, 0, c+4) -> D(c, a+3, 0) if c <= 2
E(a, 0, c+7) -> E(1, a+3, c+1)
E(a, b+1, c) -> D(a+c+1, b, 0) if 1 <= c <= 2
E(a, b+1, c+3) -> E(a+2, b, c+1)

Start: D(1, 0, 0), reached after 14 raw TM steps
\end{minted}

This initial acceleration can be simplified by composing transitions through $D$, yielding a system expressed in terms of the configuration $E$:
\begin{minted}[breaklines, fontsize=\footnotesize]{text}
Consider:
E(a, b, c) = 0^inf 10^b 1^{2a+2} E> 1^{c+1} 0^inf

E(a+1, b, 0) -> E(a, b, 1)
E(0, 0, 0) -> E(1, 0, 1)
E(0, 1, 0) -> E(1, 0, 2)
E(0, 2, 0) -> E(1, 0, 6)
E(0, b+3, 0) -> E(1, b, 4)

E(a, 0, c) -> E(1, 0, 2a+2c+1) if 1 <= c <= 2
E(a, 0, 3) -> Halt
E(a, 0, c+4) -> E(1, a+1, 2c+2) if c <= 2
E(a, 0, c+7) -> E(1, a+3, c+1)

E(a, 1, c) -> E(1, 0, 2a+2c+2) if 1 <= c <= 2
E(a, 2, c) -> E(1, 2, 2a+2c) if 1 <= c <= 2
E(a, b+3, c) -> E(1, b, 2a+2c+4) if 1 <= c <= 2
E(a, b+1, c+3) -> E(a+2, b, c+1)

Start: E(1, 0, 2), reached after 28 raw TM steps
\end{minted}

\subsection{New Discovery: \texttt{1RB1LA\_\allowbreak 1LC0RE\_\allowbreak 1LF1LD\_\allowbreak 1RE0LC\_\allowbreak 1LB1RA\_\allowbreak -{}-{}-0LA}}
\texttt{1RB1LA\_\allowbreak 1LC0RE\_\allowbreak 1LF1LD\_\allowbreak 1RE0LC\_\allowbreak 1LB1RA\_\allowbreak -{}-{}-0LA}  is a holdout in $\mathrm{BB}(6)$ for which no acceleration had previously been documented by the community. Our pipeline discovered and formally verified the following Collatz-like acceleration for this machine:
\begin{minted}[breaklines, fontsize=\footnotesize]{text}
Consider:
C(a,b,c,d) = 0^inf 1^{a+1} C> 00 1^{b+4} 01^c 1^d 011 0^inf
M(a,b,c,d) = 0^inf 1^{a+3} 01 A> 0 1^{b+4} 01^c 1^d 011 0^inf
L(a,b,d)   = 0^inf 1^{a+3} 0 1^{b+6} C> 1^{d+1} 011 0^inf
where d in {3, 6}

Identity:
C(a,b,0,6) = C(a,b+3,0,3)

C(a+2,b,c,d) -> C(a,b+3,c,d)
C(0,b,c,d) -> M(0,b,c,d)
C(1,b,c,d) -> M(1,b,c,d)

M(a,b,c+2,d) -> C(a+b+4,3,c,d)
M(a,b,1,6) -> C(a+b+4,0,3,3)
M(a,b,0,3) -> L(a,b+1,3)
M(a,b,1,3) -> L(a,b,6)

L(a,2k+1,d) -> C(a,0,k+3,d)
L(0,2k,d) -> Halt
L(1,0,6) -> C(6,0,3,3)
L(1,2k+2,d) -> C(6,3,k,d)

Start: C(4,0,0,3), reached after 162 raw TM steps
\end{minted}

This initial acceleration can be simplified by composing transitions through $M$ and $L$, yielding a system expressed in terms of the configuration $C$:
\begin{minted}[breaklines, fontsize=\footnotesize]{text}
Consider:
C(a,b,c,d) = 0^inf 1^{a+1} C> 00 1^{b+4} 01^c 1^d 011 0^inf
where d in {3, 6}

C(a,b,0,6) = C(a,b+3,0,3)

C(a+2,b,c,d) -> C(a,b+3,c,d)

C(a,b,c+2,d) -> C(a+b+4,3,c,d) if a <= 1
C(a,b,1,6) -> C(a+b+4,0,3,3) if a <= 1

C(a,2k,0,3) -> C(a,0,k+3,3) if a <= 1
C(0,2k+1,0,3) -> Halt
C(1,2k+1,0,3) -> C(6,3,k,3)

C(a,2k+1,1,3) -> C(a,0,k+3,6) if a <= 1
C(0,2k,1,3) -> Halt
C(1,0,1,3) -> C(6,0,3,3)
C(1,2k+2,1,3) -> C(6,3,k,6)

Start: C(4,0,0,3), reached after 162 raw TM steps
\end{minted}

\section{Rediscoveries}
\label[appendix]{app:rediscoveries}
Table~\ref{tab:community-overlap} compares the accelerations autonomously found and verified by our pipeline with those previously proposed by the community for the fifteen rediscovery machines.

\begin{table}[H]
\centering
\small
\setlength{\tabcolsep}{4pt}
\renewcommand{\arraystretch}{1.15}
\caption{Relation between the macro-rule systems we prove in Lean~4 and the rules previously published by the bbchallenge community (Wiki/Discord), for the 15 \emph{rediscovery} machines. Correspondences were checked by direct simulation.}
\label{tab:community-overlap}
\begin{tabular}{
  >{\raggedright\arraybackslash}p{0.15\textwidth}
  >{\raggedright\arraybackslash}p{0.32\textwidth}
  >{\centering\arraybackslash}p{0.03\textwidth}
  >{\raggedright\arraybackslash\scriptsize}p{0.42\textwidth}}
\toprule
\textbf{Class} & \textbf{Meaning} & \textbf{\#} & \textbf{Machines} \\
\midrule
\textbf{Exact}
& Same configuration families and the same transitions as the published rules (up to renaming, additive parameterization, or a one-step head phase). Our contribution is the formal closure proof.
& 2
& \texttt{1RB1LE\_1LC0RA\_0RF0LD\_1LE1LA\_1RC0LB\_---1RC}\newline
  \texttt{1RB0LD\_0RC1RB\_0RD0RA\_1LE0RD\_1LF---\_0LA1LA} (BMO\,8) \\
\addlinespace
\textbf{Equivalent, reformulated}
& Same dynamics, but our families or acceleration level differ: typically we expose an extra sweep parameter, so each published rule is a composition (or a slice) of several of ours, and the community's higher-level (exponential) formulations are not proved.
& 8
& \texttt{1RB0RF\_1RC1RF\_1LD0LE\_---1LC\_1RA1LE\_1LC1RB}\newline
  \texttt{1RB1LC\_1RC0LD\_1LA0RB\_1LB1LE\_1RF0LA\_---0RE} (BMO\,5, TNF~C)\newline
  \texttt{1RB1LC\_1LC1RD\_1LA0LB\_0RB1RE\_1RF0RA\_---0RC}\newline
  \texttt{1RB1RE\_1LC0RA\_0RD1LB\_---1RC\_1LF1RE\_0LB0LE} (BMO\,1)\newline
  \texttt{1RB0LD\_1LC0RA\_1RA1LB\_1LA1LE\_1RF0LC\_---0RE} (BMO\,5)\newline
  \texttt{1RB1RE\_1LC0RA\_1RD0LB\_1LB1RC\_1LF0RD\_---0LE} (BMO\,5, TNF~D)\newline
  \texttt{1RB3LA1LA1RA1RA\_2LB2RA---4RB1LB} (BB(2,5))\newline
  \texttt{1RB1RF\_0LC0RF\_1RD1LC\_---0LE\_0RC1LF\_1RA0LE} \\
\addlinespace
\textbf{Different decomposition}
& Genuinely different rule systems: the community has no closed system (a single partial rule, or closure conditional on an unproved number-theoretic inequality), or our proof sits at a different acceleration level (one-step CTL vs.\ macro rules).
& 3
& \texttt{1RB1LA\_1RC1RE\_1LD0RB\_1LA0LC\_0RF0RD\_0RB---}\newline
  \texttt{1RB1RF\_1LC1LF\_0RE1LD\_0LB1LD\_---1RC\_1RA0RD}\newline
  \texttt{1RB---\_0RC0RD\_1LD1RB\_0LE0LC\_1RA0LF\_1LD1LE} \\
\addlinespace
\textbf{Undetermined}
& Community rules were only posted on Discord and we were not able to reproduce.
& 2
& \texttt{1RB0LE\_1LC1RA\_1RA1LD\_0LC0LA\_0RE1RF\_0RC---}\newline
  \texttt{1RB0RB\_1LB0LC\_1RD1LC\_1RE0RA\_0RF1RA\_---1LB} \\
\bottomrule
\end{tabular}
\end{table}

\section{Candidate Machines}
\label[appendix]{app:candidates}
This section lists the 52 candidate machines identified during the initial conjecturing phase: 50 in $\mathrm{BB}(6)$ and two in $\mathrm{BB}(2,5)$. Each machine yielded two or more conjectures passing the strict 100\% empirical accuracy threshold.
Several of these machines had already been studied by the community. Some have since been resolved and are no longer holdouts.

\begin{minted}[breaklines, fontsize=\footnotesize]{text}
1RB0LD_1LC1RA_1RA1LC_1LE1LA_0RF1LD_---1RE
1RB1LD_1RC0RA_1LA1RE_0LA0LB_0RB1RF_---1RD
1RB1LF_0RC1RD_1LA1LD_1LE0RA_---1LC_0LA1LA
1RB0LD_1LC0RA_1RA1LB_1LA1LE_1RF0LC_---0RE
1RB---_1RC0LC_0RD0LE_1LE1RF_0LB0RC_0RE1RA
1RB1RD_1LC0LE_1RA0LB_1RA1RF_0RA0RC_1RE---
1RB1RC_1LC0LE_0RA1RD_1RF1LB_1LD0LA_---0RC
1RB1LC_1LA0RD_0LB1LF_1RE1LC_1LC0RB_0LA---
1RB1RD_0LC1RA_1RA1LC_0RB1LE_1RF0LD_---0RE
1RB---_1LC1RE_1LD0LC_1LE1LC_1RF0RE_1RA0RB
1RB0LF_0RC1RD_0LD0RA_1RC0RE_1RF---_1LA0LB
1RB0RE_0LC---_1RE1LD_0LE1LB_1LC0RF_1RA1LD
1RB1RE_1LC0RA_1RD0LB_1LB1RC_1LF0RD_---0LE
1RB1LA_1RC1RE_1LD0RB_1LA0LC_0RF0RD_0RB---
1RB0RB_1LB0LC_1RD1LC_1RE0RA_0RF1RA_---1LB
1RB0LF_1RC1RA_0RD0RB_1LE---_0LA1LE_1RD1LA
1RB1LE_1RC0RF_0LD---_1RF1LE_0LF1LC_1LD0RA
1RB1RE_1LC0RA_0RD1LB_---1RC_1LF1RE_0LB0LE
1RB0RE_1LC0RA_1LA1LD_1LC1LF_0LC0LB_1LE---
1RB0LC_1RC1RE_1LA0LD_0RB0RA_1RB1RF_1RD---
1RB0LB_0RC0LD_1LD1RE_0LA0RB_0RD0RF_1RA---
1RB---_0RC0RD_1LD1RB_0LE0LC_1RA0LF_1LD1LE
1RB1RD_0LC1RA_0RA1LC_1RA1LE_1RF0LD_---0RE
1RB1LE_1LC0RA_0RF0LD_1LE1LA_1RC0LB_---1RC
1RB0RF_1LC---_1RD0LC_0RE0RD_1LF1LC_1LE1RA
1RB0LA_0RC0RB_1LD1LA_1LC1RE_1RF0RD_1LC---
1RB1RC_0LA1RF_1RD0LE_1LC0RB_0RA0LC_0RD---
1RB0RE_1LC1RA_1LD0LC_1RA0LA_1RA1RF_1RD---
1RB1LC_1LC0RF_0LF1LD_0LE---_1RF1LC_1LE0RA
1RB1LE_1RC1RA_0LD1RB_1LE1LD_1RF0LA_---0RE
1RB0RE_1LC0RA_0LD1LB_1LA1LE_0LB0LF_1LD---
1RB1RF_1LC1LF_0RE1LD_0LB1LD_---1RC_1RA0RD
1RB0RF_1RC1RF_1LD0LE_---1LC_1RA1LE_1LC1RB
1RB1RF_0LC0RF_1RD1LC_---0LE_0RC1LF_1RA0LE
1RB1RB_1LC0RE_---1LD_0RA1LB_1RA1LF_0LE1LE
1RB1RC_1LC1LC_1RE0LD_1LB1RF_---1RA_0RD1RD
1RB1LA_1LC0RE_1LF1LD_1RE0LC_1LB1RA_---0LA
1RB1RA_1LC0RE_1LE0LD_0LB1LB_1RA0LF_1LD---
1RB1LF_1LC0RD_---1LA_0LA0RE_1LA1RA_1LB0LD
1RB1LC_1RC0LD_1LA0RB_1LB1LE_1RF0LA_---0RE
1RB0LF_1LC1RA_0RE0RD_---1LE_1LF1RC_1LC1LA
1RB1RE_1LC1RE_---1LD_1LE0LB_0RF0LD_1RD1RA
1RB1RF_1LC1LB_1RD1LD_1RE0RA_---0LB_0LF0RC
1RB0LC_1LA1RF_1LD1RF_1LE0LA_0RB---_0RA1RE
1RB1LC_1LC1RD_1LA0LB_0RB1RE_1RF0RA_---0RC
1RB0LE_1LC1RA_1RA1LD_0LC0LA_0RE1RF_0RC---
1RB0LD_0RC1RB_0RD0RA_1LE0RD_1LF---_0LA1LA
1RB0LF_1LC0RB_0LD1LA_0RA1LE_0LA0RA_1LA---
1RB0LC_1LA0RE_0RD1LF_1LC1LB_0LD0RB_0LA---
1RB0LC_0LC1RF_1LE1RD_0RC1RC_0LA1LA_---1RA
1RB3LA1LA1RA1RA_2LB2RA---4RB1LB
1RB2LA0RB1LB---_1LA3RA1RA4LB0RB
\end{minted}

\section{Tape Compression Algorithm}
\label[appendix]{app:tape-compression}
In this section, we present the algorithm we use to convert raw
Turing machine configurations into a compact representation based
on repeated tape symbols. The algorithm takes as input a Turing machine configuration, consisting of the head, its position, and the Turing machine tape. It detects repeated blocks (sequences of tape symbols) and replaces consecutive repetitions  with the block and its repetition count, while preserving the head marker and infinite blank regions.

For example, the raw Turing machine configuration, written in directed head notation:
$$0^\infty\,01010101\,\mathrm{E}{>}\,1111111\,0^\infty$$
is compressed into:
$$0^\infty\ 01^4\ \mathrm{E}{>}\ 1^7\ 0^\infty$$
where $0^\infty$ denotes an infinite sequence of blank symbols and $\mathrm{E}{>}$ indicates that the machine is in state E with the head reading the symbol immediately to its right.
The compressed representation is exact: expanding each block according to its repetition count recovers the original configuration.

We compress the finite sequences of symbols on either side of the head independently, keeping the head marker and infinite blank regions unchanged. Algorithm~\ref{alg:compress_sequence} performs this scan by alternating between two steps: identifying a repeating block and adjusting its boundary. Each side is scanned from its outer boundary toward the head. At a given position, we identify repeating sequences of symbols starting there and use a simplicity score to choose among them.
We then check whether taking one fewer copy of the chosen block would allow a longer repetition to start at the resulting boundary. We append the chosen block and repetition count to the output and continue with the remaining symbols. The following two steps describe this procedure for a sequence written in its scanning order.

The first step, implemented in Algorithm~\ref{alg:best_block},
identifies the next block. Let $s$ denote the sequence of symbols
still to be processed. For each candidate block $b$ starting at
the beginning of $s$, we count its maximum number $n$ of consecutive
copies. Among candidates with $n\geq 2$,
we maximize the simplicity $\operatorname{score}(b,n)=\frac{n}{|b|},$
where $|b|$ is the number of symbols in $b$. This favors short, frequently repeated blocks when several decompositions are possible: $01010101=01^4=0101^2$, but the respective scores are $2$ and $1/2$. Ties favor shorter blocks. If no candidate repeats, we select the first symbol with count $1$. Candidate lengths range from $1$ to $\lfloor\min(|s|,K)/2\rfloor$, where $K$ is a search parameter (default $16$, allowing blocks of length at most $8$).
This step returns the candidate block $b$ and its repetition count $n$ to the main procedure.

\begin{algorithm}[H]
\caption{Block selection using the simplicity score}
\label{alg:best_block}
\begin{algorithmic}[1]
\Require Nonempty sequence of symbols $s$; integer $K\geq 2$
\Ensure Block and repetition count $(b,n)$
\Function{BestBlock}{$s,K$}
    \State $\mathrm{bestBlock}\gets(s[0],1)$;
        $\mathrm{bestScore}\gets 0$
    \For{$\mathrm{blockLength}=1,\ldots,
        \lfloor\min(|s|,K)/2\rfloor$}
        \State $b\gets s[0:\mathrm{blockLength}]$; $n\gets 1$
        \While{$(n+1)\mathrm{blockLength}\leq |s|$ \textbf{and}
            $s[n\mathrm{blockLength}:(n+1)\mathrm{blockLength}]=b$}
            \State $n\gets n+1$
        \EndWhile
        \If{$n\geq 2$ \textbf{and}
            $\operatorname{score}(b,n)>\mathrm{bestScore}$}
            \State $\mathrm{bestBlock}\gets(b,n)$
            \State $\mathrm{bestScore}\gets\operatorname{score}(b,n)$
        \EndIf
    \EndFor
    \State \Return $\mathrm{bestBlock}$
\EndFunction
\end{algorithmic}
\end{algorithm}

\begin{algorithm}[H]
\caption{Sequence compression with boundary adjustment}
\label{alg:compress_sequence}
\begin{algorithmic}[1]
\Require Finite sequence of symbols $s$; integer $K\geq 2$
\Ensure List $\mathrm{compressed}$ of blocks and repetition counts
\State $\mathrm{compressed}\gets[\,]$
\While{$|s|>0$}
    \State $(b,n)\gets\Call{BestBlock}{s,K}$
    \State $\mathrm{fullLength}\gets n|b|$
    \If{$n\geq 2$ \textbf{and} $\mathrm{fullLength}<|s|$}
        \State $\mathrm{shortLength}\gets\mathrm{fullLength}-|b|$

        \State $(b_{\mathrm{next}},n_{\mathrm{next}})
            \gets\Call{BestBlock}{s[\mathrm{fullLength}:],K}$
        \State $\mathrm{fullCoverage}\gets
            \mathrm{fullLength}+n_{\mathrm{next}}|b_{\mathrm{next}}|$

        \State $(b_{\mathrm{next}},n_{\mathrm{next}})
            \gets\Call{BestBlock}{s[\mathrm{shortLength}:],K}$
        \State $\mathrm{shortCoverage}\gets
            \mathrm{shortLength}+n_{\mathrm{next}}|b_{\mathrm{next}}|$

        \If{$\mathrm{shortCoverage}>\mathrm{fullCoverage}$}
            \State $n\gets n-1$
        \EndIf
    \EndIf
    \State Append $(b,n)$ to $\mathrm{compressed}$
    \State $s\gets s[n|b|:]$
\EndWhile
\State \Return $\mathrm{compressed}$
\end{algorithmic}
\end{algorithm}

Before committing to this candidate, the main procedure checks whether its boundary should be adjusted. Taking all $n$ copies can consume the start of another repetition, so we compare keeping $n$ copies with keeping $n-1$. For each choice, we call Algorithm~\ref{alg:best_block} on the remaining symbols and measure the total number of symbols covered by the current and next blocks. We take one fewer copy only if this strictly increases that total. This check is skipped when $n=1$ or the selected block reaches the end of the sequence. The next block is examined only to guide this decision, it is not yet added to the output.

For example, on input $1111110101010$, block selection gives $1^6$.
The next block would be $01^3$, covering $6+6=12$ symbols in total
and leaving a final $0$:
$$
\underbrace{1^6\ 01^3}_{12\text{ symbols}}0
$$
Taking only five initial $1$s instead exposes $10101010=10^4$, giving a total coverage of $5+8=13$:
$$
\underbrace{1^5\ 10^4}_{13\text{ symbols}}.
$$

After this adjustment, we append the current block and its final repetition count to the output, remove the corresponding symbols from the beginning of $s$, and repeat both steps on the remaining sequence. The procedure terminates when $s$ is empty. Applying it to both sides of the head and restoring their original orientation yields the compressed configuration.


\section{Prompt for Stage 2}
\label[appendix]{app:prompt_proving}

\begin{minted}[breaklines=true]{markdown}
[/goal](slashCommand;goal) file: busy_beavers/suspects/suspect_32/potential_proof.lean

# TASK: Synthesize Closed Pattern System & Formal Lean 4 Proof for suspect_32

You are an expert formal verification and automated theorem proving agent.
Your goal is to write a complete, machine-checked Lean 4 proof in
`busy_beavers/suspects/suspect_32/potential_proof.lean`
establishing that `suspect_32_tm` is cryptid (`IsCryptid suspect_32_tm`),
resolving `suspect_32_behavior` with **zero `sorry`**.

---

## 1. Objectives & Design Constraints

1. **Minimal Pattern System**:
   * Synthesize a closed inductive invariant with as few tape patterns and
     transition rules as possible (target: **< 15 rules**).
   * Avoid redundant or overlapping patterns. Keep canonical configurations
     clean and parameter spaces well-defined.

2. **Gold Standard Reference (MUST READ)**:
   * You can look at proved positive examples in the codebase for reference
     (`busy_beavers/suspects/suspect_09/potential_proof.lean` and
     `proved.md`). These serve as the exact reference gold-standard for how
     to formulate and prove an `IsCryptid` closed pattern system in Lean 4
     with 0 `sorry`.

3. **Execution Integrity**:
   * The proof must compile cleanly down to the Lean 4 kernel with 0 errors,
     0 warnings, and 0 `sorry`.

---

## 2. Adversarial Verifier Compliance (Must Pass)

Your generated `potential_proof.lean` will be independently audited by a
strict verification agent executing the following prompt. **Ensure your proof
strictly adheres to all constraints below:**

```text
# TASK: Formally Verify Busy Beaver Candidate Proofs with Anti-Cheat Auditing

System Role: Formal verification auditor for Lean 4 candidate proofs in
busy_beavers/suspects/<SUSPECT_NAME>/.

### 1. Strict Verification Principle
Compilation Over Text Matching: A proof is valid ONLY if it elaborates to the
Lean 4 kernel without vacuous hypotheses or gaps. Superficial AST/diff
matching is rejected.

### 2. Anti-Cheating & Non-Vacuity Auditing (Stellar BB6 Protocol)
Every proof must pass all 5 verification gates:

- Vector A (Satisfiable Hypotheses): Preconditions on macro-step rules must
  be mutually satisfiable over Nat and logically consistent. Any premise
  simplifying to False (omega/cases contra exploits) results in immediate
  DISQUALIFICATION.
- Vector B (Structural Non-Empty Invariants): The pattern set ps in
  rules_closed / cryptid must be non-empty (ps != []) and evaluate to the
  canonical configuration family (Config_*). Trivial 1-step unrolling or
  universal catch-all patterns are DISQUALIFIED.
- Vector C (Genuine Blank-Tape Reachability): The entry theorem
  (init_reaches_*) must prove init [] executes a strictly positive number of
  concrete machine steps (|-+ / execute) to reach the canonical Config_*
  family. Zero-step bypasses (cfg_0 = init []) or disconnected dummy patterns
  are DISQUALIFIED.
- Vector D (Exhaustive Parameter Coverage): Macro-step rules must form a
  total, exhaustive partition of N^k for all valid variable assignments in
  ps. Proofs relying on Nat truncated subtraction underflow (a - b = 0 when
  a < b) or leaving unhandled parameter gaps are DISQUALIFIED.
- Vector E (Rule Proof Integrity): Every rule_* must be proved bottom-up
  from the primitive transition function step with strictly positive progress
  (|-+) and zero unproved axioms (no Lean.ofReduceBool, no foreign axioms).
  Any subtraction within parameter bounds must have explicit inequality
  proofs preventing underflow.

### 3. Verification Pipeline Checks
1. Kernel compilation: 0 errors, 0 warnings, 0 sorry.
2. Axiom audit (#print axioms): Only propext, Classical.choice, Quot.sound.
3. Comparator check: All declarations from problem.lean must match bit-for-bit.
4. Kernel replay: Full export via lean4export validated by independent checker.
\end{minted}

\end{document}